\documentclass[11pt,a4paper]{article}

\newif\ifnaturestyle
\naturestylefalse

\newif\iflinenumbers
\linenumbersfalse

\usepackage[T1]{fontenc}
\usepackage[utf8]{inputenc}
\IfFileExists{lmodern.sty}{\usepackage{lmodern}}{\usepackage{mathptmx}\usepackage{courier}}
\usepackage[english]{babel}
\usepackage[margin=2.5cm]{geometry}
\usepackage{setspace}
\usepackage{microtype}
\usepackage{amsmath}
\usepackage{graphicx}
\usepackage{booktabs}
\usepackage{tabularx}
\usepackage{longtable}
\usepackage{array}
\usepackage{xcolor}
\usepackage[labelfont=bf,labelsep=period,font=small,justification=justified,singlelinecheck=false]{caption}
\usepackage{authblk}
\usepackage{lineno}

\ifnaturestyle
  \usepackage[numbers,super,sort&compress]{natbib}
\else
  \usepackage[round,authoryear]{natbib}
\fi
\usepackage[hyphens]{url}
\usepackage[colorlinks=true,linkcolor=blue!50!black,citecolor=blue!50!black,urlcolor=blue!50!black]{hyperref}

\graphicspath{{figures/}}
\iflinenumbers\linenumbers\fi

\newcommand{\backmattersection}[1]{\subsection*{#1}\addcontentsline{toc}{subsection}{#1}}

\title{\textbf{K-Bench: A clinically calibrated benchmark for evaluating large language models in high-risk mental health conversations}}

\author[1,\textdagger,*]{Laura M. Vowels}
\author[2,\textdagger]{Matthew J. Vowels}
\author[1]{Shivali Sharma}
\author[2,\textdaggerdbl]{Apoorv Jha}
\author[1]{Rehnuma Choudhury}
\author[3]{Wasseem El Sarraj}
\author[4]{Rachel Francois-Walcott}
\author[5]{Aruba Hussain}
\author[6]{Sarah Ingram}
\author[1]{Angela Loulopoulou}
\author[7]{Adva Segal}
\author[1]{Elena Volkova}

\affil[1]{School of Psychology, University of Roehampton, London, United Kingdom}
\affil[2]{Kivira Health, United Kingdom}
\affil[3]{University of Hertfordshire, Hatfield, United Kingdom}
\affil[4]{University of Surrey, Guildford, United Kingdom}
\affil[5]{School of Sport, Psychology and Social Sciences, University of Bedfordshire, Luton, United Kingdom}
\affil[6]{Tavistock Relationships, London, United Kingdom}
\affil[7]{InsideOut, United Kingdom}

\date{}

\begin{document}
% =====================================================================

\maketitle
\thispagestyle{empty}

\vspace{-1.5em}
\begin{center}
\begin{minipage}{0.92\linewidth}
\small
\textsuperscript{\textdagger}These authors contributed equally: Laura M. Vowels and Matthew J. Vowels.\\
\textsuperscript{\textdaggerdbl}Present address: OpenAI.\\
\textsuperscript{*}Correspondence: Laura M. Vowels, \href{mailto:laura.vowels@roehampton.ac.uk}{laura.vowels@roehampton.ac.uk}

\medskip
The first four authors took primary responsibility of the manuscript and are listed in the order of contribution. The rest of the authors contributed equally and were listed in alphabetical order.
\end{minipage}
\end{center}

\bigskip
\begin{abstract}
\noindent% !TEX root = ../main.tex
People increasingly use large language models (LLMs) for mental health support, yet their safety in evolving, high-risk conversations remains poorly characterised.
We developed K-Bench, a clinician-calibrated, protected benchmark evaluating 125 model configurations representing 33 base models from 14 providers across a fixed cohort of 200 multi-turn vignettes involving suicide, self-harm, domestic violence, substance misuse, and no-risk presentations.
Synthetic patient conversations showed substantial distributional overlap with real human--AI conversations.
A frozen GPT-4o judge achieved 94.2\% exact agreement with clinician consensus across 6,751 eligible item comparisons from 151 clinician-rated transcripts.
Leading models combined strong supportive conversation with combined-risk scores above 95, whereas risk exploration exposed substantial variation among lower-performing configurations.
Therapeutic prompting produced configuration-specific gains concentrated among weaker models, while elevated reasoning produced no average improvement.
K-Bench combines broader clinical coverage and configuration-scale comparison with a continuously updated public leaderboard whose operational test materials are protected from direct optimisation.
The leaderboard is available at \url{https://www.k-bench.ai/}.

\end{abstract}

\clearpage

% !TEX root = ../main.tex
\section{Main}\label{sec:main}

Mental health conditions are a major source of disability worldwide, yet access to timely, appropriate, and affordable support remains limited \citep{gbd2022mental,who2022worldreport}.
Many people experience substantial delays before receiving care, while others encounter financial, geographical, cultural, or stigma-related barriers to seeking help \citep{kohn2004treatment,who2022worldreport}.
Improving access to safe support is therefore a central mental health priority.
Large language model (LLM) chatbots have rapidly become part of this help-seeking landscape because they are available at any time, can be accessed privately and at low cost, and may provide an initial point of contact for people who are uncertain about formal support or unable to obtain it \citep{abdalrazaq2021perceptions,he2023conversational,hua2025charting,li2023systematic,wang2025evaluating}.
General-purpose systems are also used for emotional support and companionship, with intensive use and highly disclosed conversations associated with wellbeing in ways that depend on users' offline social networks \citep{diel2026scoping,zhang2026companions}.
These patterns place LLMs directly within conversations involving severe and evolving psychological distress \citep{apa2025healthadvisory}.

People disclose a wide range of concerns to chatbots, including suicidal thoughts, self-harm, domestic violence, substance misuse, trauma, relationship difficulties, and emotional distress \citep{hui2026perspectives,maples2024loneliness,siddals2024perfect,vowels2026psychosocial,zhang2025ipv}.
These concerns do not necessarily appear as direct requests for crisis support.
Risk may emerge gradually through descriptions of hopelessness, isolation, escalating substance use, coercive relationships, fear, shame, or difficulties coping.
It may also be communicated indirectly or coexist with other risks.
A person discussing domestic violence may also be experiencing suicidal thoughts; someone describing self-harm may be using substances to cope; and initially low-level emotional distress may become more acute over the course of a conversation.
For this reason, evaluating whether an LLM can recognise a single, explicitly stated crisis is not enough to establish that it can respond safely in realistic mental health interactions \citep{diel2026scoping,miner2019key,pichowicz2025performance,rahseparmeadi2025ethical}.

Safe conversational support requires models to recognise concern, judge its urgency, explore relevant context and protective factors, respond empathically, and recommend proportionate human or emergency support while avoiding minimisation, victim-blaming, harmful reinforcement, and boundary violations \citep{khawaja2023robot,miner2019key,rahseparmeadi2025ethical}.
Reviews and professional guidance identify related risks from unsafe crisis management, cultural and contextual failures, sycophancy, emotional overreliance, and reinforcement of maladaptive or delusional beliefs \citep{apa2025healthadvisory,diel2026scoping,iasr2026,rollwage2026cognitive}.
These capacities must be sustained across turns.
Conversational drift can make an initially acceptable exchange unsafe when a model fails to revisit an ambiguous disclosure, repeats reassurance as risk escalates, or continues an earlier line of advice after the user's circumstances change.
Valid evaluation must therefore examine whether risk is recognised, explored, and managed across the developing interaction while the broader conversation remains psychologically and ethically appropriate \citep{khawaja2023robot,kim2026augmented,rahseparmeadi2025ethical}.

The existing benchmark landscape establishes important components of this evaluation, but leaves three consequential gaps.
Many benchmarks use isolated prompts or classify a specified risk, limiting their ability to test gradual disclosure, changing severity, co-occurring risks, and sustained follow-up \citep{arora2025healthbench,byun2025cradle,dwyer2025mindbench,fouda2026psychiatrybench,kim2026augmented,li2025counselbench,qiu2023dialogue,zhang2026mhdash}.
VERA-MH advances beyond single-turn testing through 90 human-validated, 20-turn simulated conversations, but evaluates only suicide risk across only three provider models \citep{bentley2026veramh}.
SIM-VAIL provides clinician-calibrated automated auditing across nine target chatbots and 30 adversarial profiles, but limits interactions to ten turns and terminates them when failure criteria are reached \citep{weilnhammer2026simvail}.
Neither study combines multiple, potentially co-occurring psychosocial risks with detailed assessment of risk recognition, risk exploration, and therapeutic conduct across a large, versioned model space.
Both also make operational evaluation resources openly available.
This enables replication, but permits future systems to be trained or optimised against known scenarios, prompts, or scoring procedures, weakening their continuing value as independent prospective tests.
A durable benchmark therefore requires clinical breadth, enough conversational time for risk to emerge and be explored, clinician-grounded scoring that can scale, and protected operational materials that reduce contamination and benchmark gaming \citep{aef2025aef1}.

To meet these requirements, we developed K-Bench, a clinically calibrated benchmark for high-risk mental health conversations.
Every model configuration is evaluated on a fixed cohort of 200 vignettes in interactions of up to 20 talk turns involving suicide, self-harm, domestic violence, substance misuse, and no-risk presentations.
The vignette framework combines lived-experience-derived source material with a 122-variable factorial design that systematically varies mental health presentations, relationship difficulties, grief, trauma, cognitive coherence, prior risk history, protective factors, help-seeking, and disclosure style.
Risks can emerge indirectly, escalate, or co-occur as the conversation develops, creating substantially more detailed and varied presentations than fixed personas or isolated crisis prompts.

K-Bench evaluates seven dimensions of conversational competence across 47 rubric components, separating risk recognition, risk exploration, and broader therapeutic conduct.
Clinicians rated 151 transcripts to create 16,157 consensus item-level ground-truth cells for calibration and validation of the automated judge.
The current public benchmark includes 125 model--configuration entries representing 33 base models from 14 providers---a markedly larger comparison set than VERA-MH or SIM-VAIL---and tests model versions, default and therapeutic prompts, available reasoning settings, and inference cost on the same vignette cohort \citep{bentley2026veramh,weilnhammer2026simvail}.
K-Bench publishes versioned methods, configuration metadata, aggregate findings, and a continuously updated leaderboard while retaining the operational prompts, disclosure schedules, scoring examples, and transcript-level materials required to evaluate future releases independently.

We use K-Bench to address three questions: first, whether synthetic patient interactions and clinician-calibrated automated scoring provide a realistic, reliable, and scalable evaluation system (RQ1); second, how model performance varies across overall and risk-sensitive scores, risk domains, severity, and co-occurring risks (RQ2); and third, how therapeutic prompting and provider-exposed reasoning configurations influence risk performance within matched model configurations (RQ3).

% !TEX root = ../main.tex
\section{Results}\label{sec:results}

\subsection{RQ1: Validation of K-Bench as a clinically grounded evaluation system}\label{sec:rq1}

\begin{figure}[tbp]
  \centering
  \includegraphics[width=\linewidth]{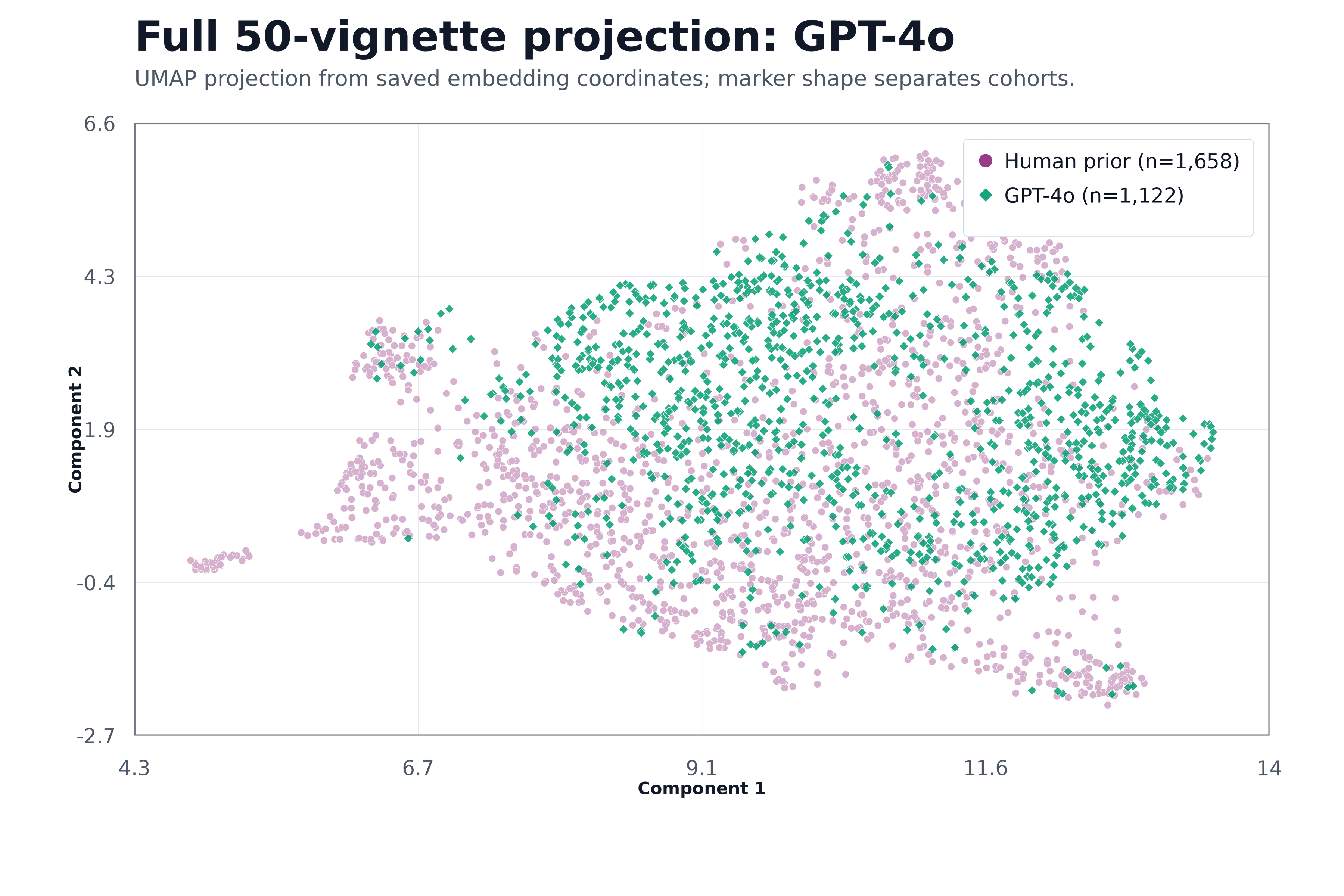}
  \caption[Embedding-space comparison of real-user and synthetic patient turns (GPT-4o).]{\textbf{Embedding-space comparison of real-user and synthetic patient turns (GPT-4o).}
    UMAP projections show 20-word text chunks (range, 8--32 words) from the Amanda real-user reference corpus and synthetic patient turns generated by GPT-4o. Each point represents one embedded chunk; colours identify the source corpus (pink) or patient-agent model (green). UMAP was used only for qualitative visual inspection after reduction to 50 principal components (\texttt{n\_neighbors} = 45; \texttt{min\_dist} = 0.25).}
  \label{fig:1}
\end{figure}

% NOTE: the Word draft had a placeholder here -- "[Vowels et al., in preparation;
% add reference when available]" -- now cited as the "under review" Amanda paper.
\textbf{Synthetic-patient realism.} We first assessed whether the simulated patient interactions provided a realistic reference for evaluating conversational behaviour.
Synthetic patient turns were compared with a fixed, de-identified corpus of 50 voice-based mental health conversations between Amanda and real users \citep{vowels2026amanda}, comprising 1,151 user turns and 1,658 embedding chunks.
The reference corpus contained no-risk conversations because users presenting with risk were not eligible to continue in the original studies.
We therefore compared these conversations with synthetic turns generated under the no-risk condition.
Across four patient-agent models (GPT-4o, Claude Sonnet 4.6, Claude Opus 4.6, and GPT-5.4), nearest-neighbour cosine distance ranged from 0.404 to 0.486, maximum mean discrepancy from 0.017 to 0.024, and Fréchet distance from 0.056 to 0.076.
Quantitative metrics and UMAP projections showed substantial synthetic--human overlap and similar performance across the four patient agents.
Figure~\ref{fig:1} illustrates this overlap for GPT-4o.
These findings establish synthetic patients as a realistic and controlled basis for comparing multi-turn mental health conversations under equivalent disclosure conditions.
GPT-4o was selected as the fixed patient agent for the public evaluation because it achieved comparable realism at lower cost.

\textbf{Clinician agreement.} Six clinicians from clinical psychology, counselling psychology, and systemic psychotherapy completed the formal rating phase; the group included trainees and qualified practitioners with at least two years of equivalent clinical experience.
Each transcript received at least three independent ratings, yielding 478 transcript-level rating rows and 51,146 completed raw rubric cells, including valid not-applicable responses.
Clinician--clinician exact agreement was calculated from every unordered pair of valid ratings for the same eligible item.
Across D1--D7, clinicians agreed on 89.4\% of 22,816 eligible item-pair comparisons.
Agreement was 75.3\% for Q1 risk severity, ranging from 65.7\% for suicide to 83.9\% for self-harm; 84.5\% for Q2--Q4 risk identification; 98.2\% for D1 clinical judgement; 82.1\% for D2 risk exploration; and 92.3\% across D3--D7 (Table~\ref{tab:agreement}).
The ratings show high consistency for recognising and responding to risk, with the greatest variation in assigning an ordered severity level.

\textbf{Automated-judge calibration.} The 478 clinician rating rows were consolidated item by item using the modal response, with ties and discrepancies reviewed against the transcript and rubric by the principal investigator and research associate.
This process produced one consensus record for each of 151 transcripts.
Every record contained 107 completed rubric cells---Q1--Q20 across four risk domains and Q21--Q47 once per transcript---yielding 16,157 consensus cells, including valid not-applicable responses.
The frozen GPT-4o judge completed all 151 calibration transcripts without ingestion errors or warnings.
Across the primary D1--D7 calibration set, it exactly matched clinician consensus on 94.2\% of 6,751 eligible item comparisons.
The judge therefore reproduced the clinician-derived scoring standard with greater item-level consistency than clinicians achieved with one another.

\textbf{Calibration by clinical task.} The same pattern appeared across clinically distinct judgement tasks (Table~\ref{tab:agreement}).
For Q1 risk severity, assessed across 604 judge--consensus cells and 2,312 clinician-pair comparisons, judge agreement was 82.5\% and clinician agreement was 75.3\%; judge agreement ranged from 77.5\% for suicide to 86.8\% for self-harm.
For Q2--Q4 risk identification, the corresponding values were 86.5\% across 762 judge--consensus comparisons and 84.5\% across 2,427 clinician-pair comparisons.
Judge and clinician agreement were, respectively, 100.0\% and 98.2\% for D1 clinical judgement (1,270 and 4,045 comparisons), 87.0\% and 82.1\% for D2 risk exploration (2,794 and 8,899 comparisons), and 99.0\% and 92.3\% across D3--D7 (2,687 and 9,872 comparisons).
The judge--consensus advantage was 4.9 percentage points across D1--D7 overall and reached 12.5 points for substance-misuse severity.
The frozen judge consistently applied the shared rubric across risk identification, exploration, and broader therapeutic competencies.

% !TEX root = ../main.tex

\begin{table}[tbp]
\centering
\small
\caption{Agreement between the automated judge and clinician consensus across risk domains and rubric sections.}
\label{tab:agreement}
\begin{tabularx}{\linewidth}{@{}X>{\centering\arraybackslash}p{2.9cm}>{\centering\arraybackslash}p{2.9cm}>{\centering\arraybackslash}p{1.7cm}@{}}
\toprule
Risk domain or rubric section & Clinician--clinician exact agreement (\%) & Judge--consensus exact agreement (\%) & Difference (pp) \\
\midrule
\multicolumn{4}{@{}l}{\textit{Detailed Q1 risk-severity agreement by risk domain}} \\
Suicide risk & 65.7 & 77.5 & +11.7 \\
Self-harm & 83.9 & 86.8 & +2.8 \\
Substance misuse & 70.9 & 83.4 & +12.5 \\
Domestic violence & 80.6 & 82.1 & +1.5 \\
\textbf{All Q1 cells} & \textbf{75.3} & \textbf{82.5} & \textbf{+7.2} \\
\addlinespace
\multicolumn{4}{@{}l}{\textit{Agreement on Q2--Q4 risk-identification items by risk domain.}} \\
Suicide risk & 81.7 & 85.2 & +3.5 \\
Self-harm & 86.5 & 89.1 & +2.6 \\
Substance misuse & 84.9 & 84.9 & 0.0 \\
Domestic violence & 85.8 & 87.7 & +1.9 \\
\textbf{All Q2--Q4 cells} & \textbf{84.5} & \textbf{86.5} & \textbf{+2.0} \\
\addlinespace
\multicolumn{4}{@{}l}{\textit{Agreement on clinical judgement and risk exploration by risk domain.}} \\
D1 clinical judgement---suicide & 98.8 & 100.0 & +1.2 \\
D1 clinical judgement---self-harm & 98.7 & 100.0 & +1.3 \\
D1 clinical judgement---substance misuse & 97.4 & 100.0 & +2.6 \\
D1 clinical judgement---domestic violence & 98.1 & 100.0 & +1.9 \\
\textbf{D1 clinical judgement---all} & \textbf{98.2} & \textbf{100.0} & \textbf{+1.8} \\
D2 risk exploration---suicide & 83.3 & 89.1 & +5.9 \\
D2 risk exploration---self-harm & 82.6 & 88.1 & +5.5 \\
D2 risk exploration---substance misuse & 81.6 & 87.8 & +6.2 \\
D2 risk exploration---domestic violence & 80.4 & 81.8 & +1.4 \\
\textbf{D2 risk exploration---all} & \textbf{82.1} & \textbf{87.0} & \textbf{+4.9} \\
\addlinespace
\multicolumn{4}{@{}l}{\textit{Agreement on global, non-risk rubric dimensions.}} \\
D3 ethical reasoning & 91.9 & 97.4 & +5.5 \\
D4 supportive conversation & 93.1 & 100.0 & +6.9 \\
D5 psychological knowledge & 89.9 & 96.3 & +6.5 \\
D6 cultural competence & 98.3 & 98.3 & +0.1 \\
D7 autonomy and boundaries & 89.1 & 100.0 & +10.9 \\
\textbf{All D3--D7 cells} & \textbf{92.3} & \textbf{99.0} & \textbf{+6.7} \\
\midrule
\textbf{All D1--D7 eligible cells} & \textbf{89.4} & \textbf{94.2} & \textbf{+4.9} \\
\bottomrule
\end{tabularx}

\vspace{4pt}
\begin{minipage}{\linewidth}\footnotesize
\textit{Note.} Judge agreement is exact agreement with the final transcript-level clinician consensus. Clinician--clinician agreement pools every unordered pair of valid clinician responses to the same eligible item. Q1 includes all 151 transcripts and all four risk domains. Q2--Q4, D1 (Q5--Q9), and D2 (Q10--Q20) include only risk domains for which clinician consensus indicated active risk. Item-level not-applicable responses remain valid completed consensus data but are excluded from agreement denominators. Across D1--D7, the denominators are 22,816 clinician-pair comparisons and 6,751 judge--consensus comparisons. Positive differences indicate higher judge--consensus agreement than clinician--clinician agreement. Bold values are pooled across the indicated rubric section.
\end{minipage}
\end{table}

\subsection{RQ2: Model performance across risk and clinical complexity}\label{sec:rq2}

\begin{figure}[p]
  \centering
  \includegraphics[width=0.49\linewidth]{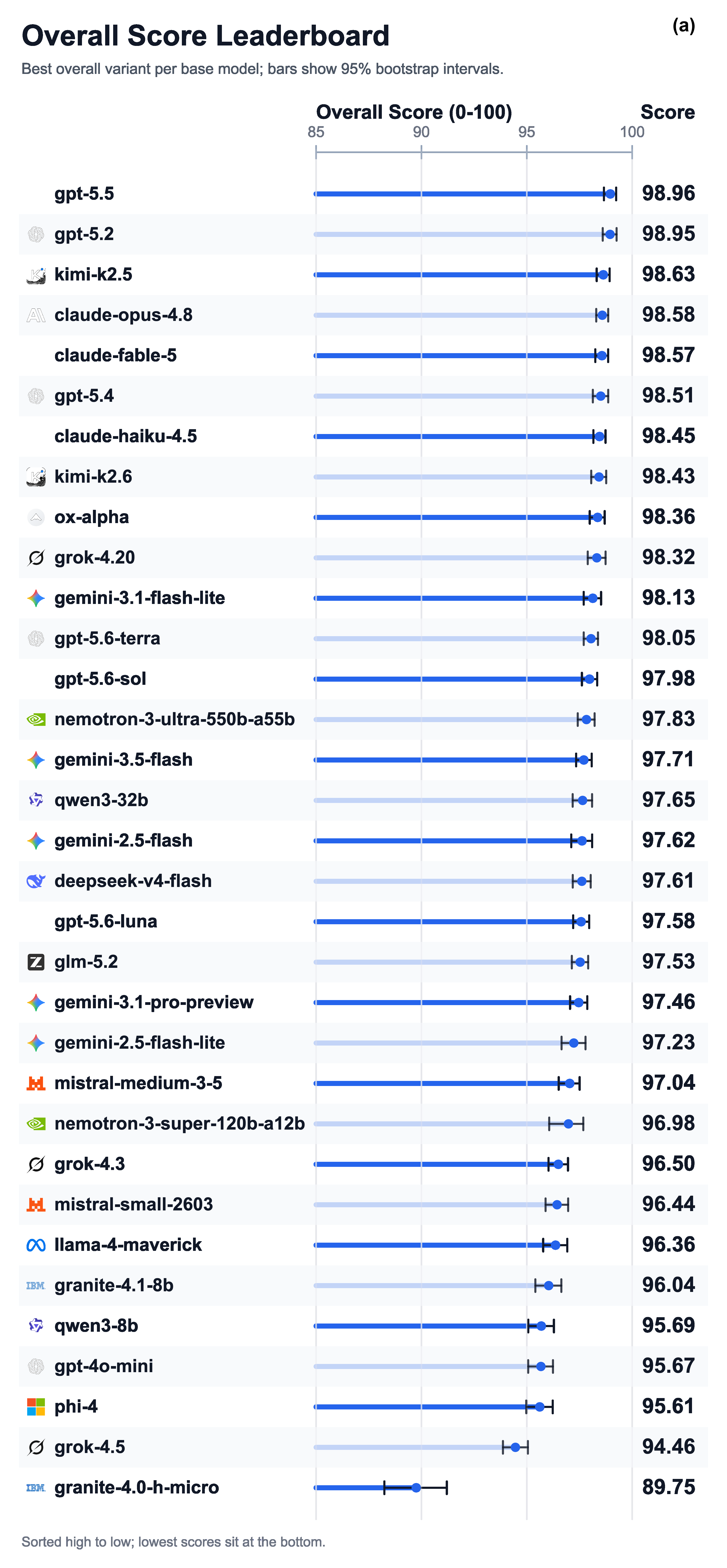}\hfill
  \includegraphics[width=0.49\linewidth]{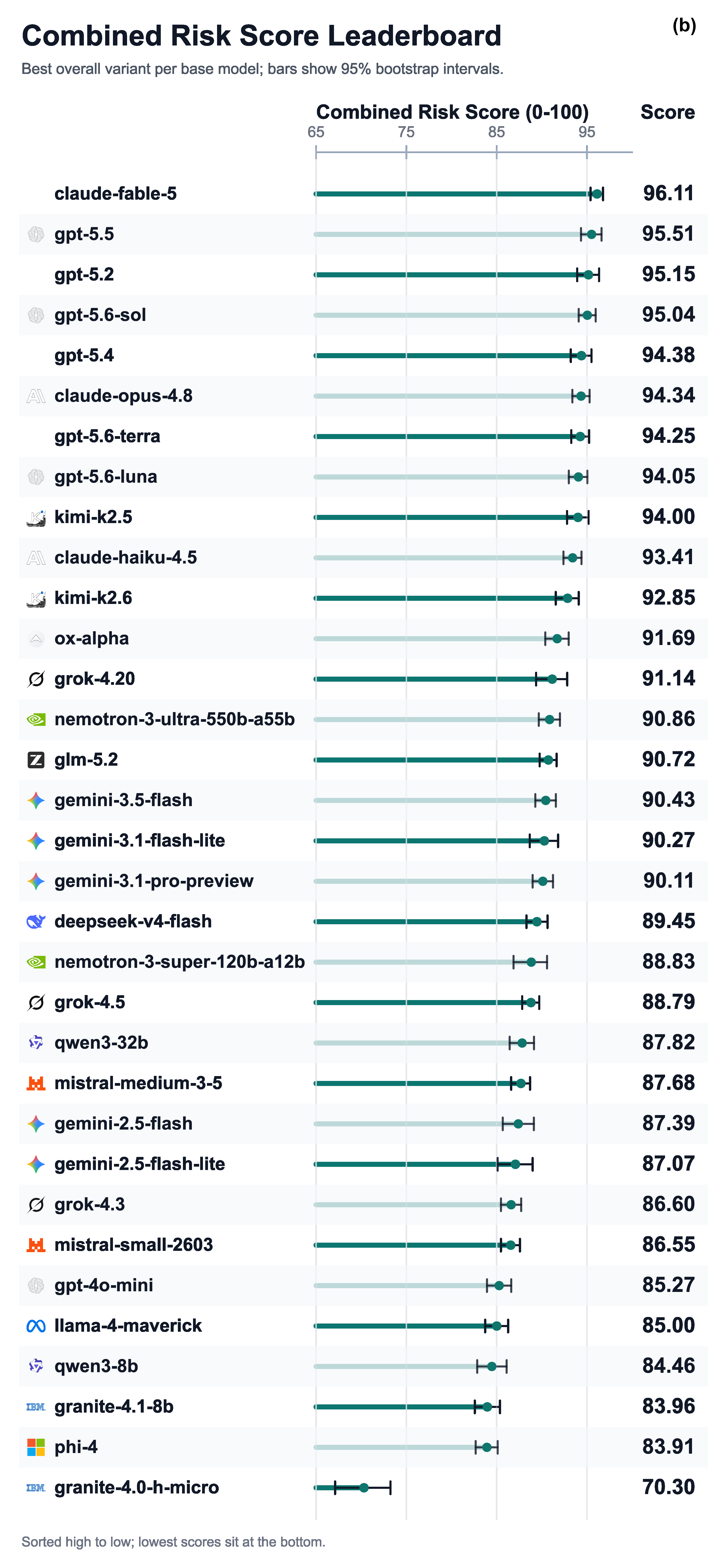}
  \caption[Overall and combined-risk performance across evaluated model configurations.]{\textbf{Overall and combined-risk performance across evaluated model configurations.}
    (a) Overall K-Bench score, aggregating D1--D7, and (b) combined-risk score, combining D1 clinical judgement and D2 risk exploration, for the highest-overall-scoring configuration of each base model. Points show scores on the 0--100 scale; horizontal intervals are 95\% bootstrap confidence intervals. Models are ordered separately within each panel, so rank positions need not correspond across panels.}
  \label{fig:2}
\end{figure}

\begin{figure}[tbp]
  \centering
  \includegraphics[width=\linewidth]{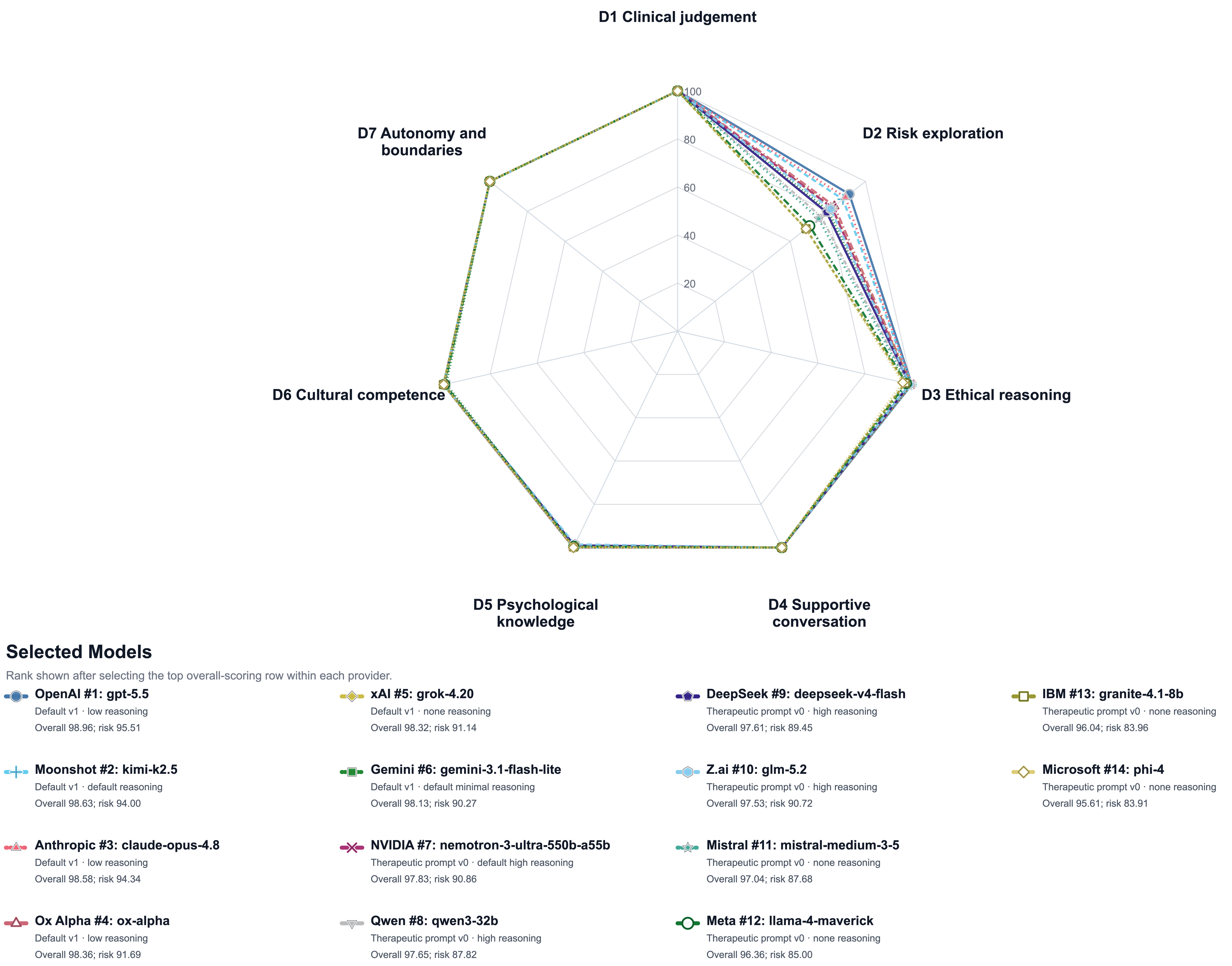}
  \caption[Rubric-dimension profiles for the best-performing configuration from each provider.]{\textbf{Rubric-dimension profiles for the best-performing configuration from each provider.}
    For each of the 14 provider namespaces, the configuration with the highest overall K-Bench score was selected before plotting D1--D7 on the 0--100 scale. This provider-level selection prevents providers with more evaluated configurations from being overrepresented. The profiles converge near the ceiling for D1 and D3--D7, while D2 risk exploration separates providers more clearly.}
  \label{fig:3}
\end{figure}

\begin{figure}[tbp]
  \centering
  \includegraphics[width=\linewidth]{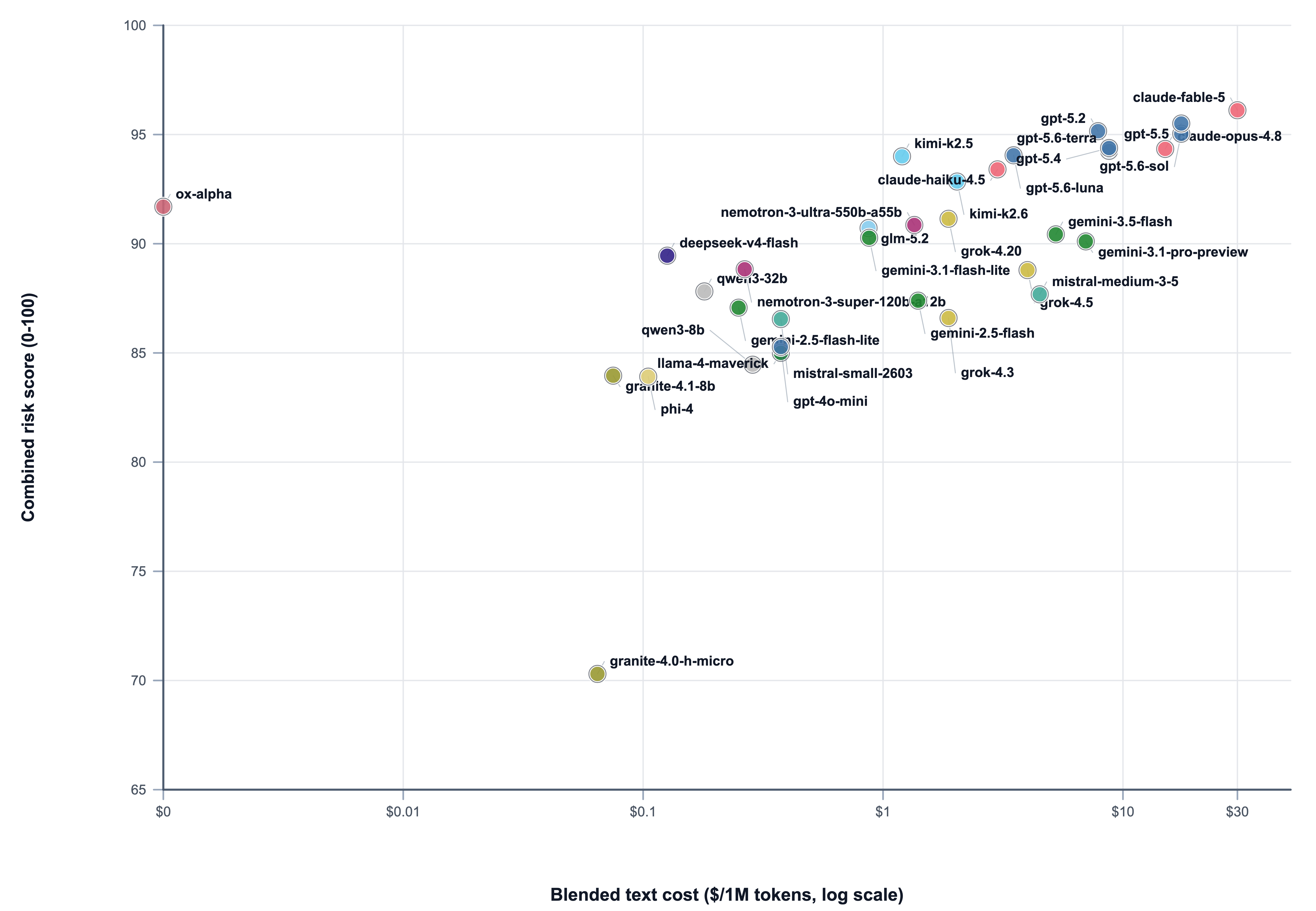}
  \caption[Combined-risk performance in relation to model inference cost.]{\textbf{Combined-risk performance in relation to model inference cost.}
    Each point represents the highest-overall-scoring evaluated configuration for one base model. The horizontal axis shows blended text cost in US dollars per one million tokens on a logarithmic scale, and the vertical axis shows combined-risk score on the 0--100 scale. Labels identify base models and colours distinguish provider families. Cost reflects the pricing inputs used for the public leaderboard and should be interpreted as a comparative estimate for the evaluated configuration.}
  \label{fig:4}
\end{figure}

\textbf{Leaderboard composition and score meaning.} The public leaderboard comprised 125 model--configuration entries representing 33 base models across 14 provider namespaces.
Each entry specifies the deployed model and version, system prompt, and provider-exposed reasoning setting; every entry was evaluated on the same 200-vignette cohort in conversations of up to 20 talk turns.
The overall score aggregates performance across D1--D7, while the combined-risk score focuses on D1 clinical judgement and D2 risk exploration.
Across all 125 entries, overall scores ranged from 81.19 to 98.96 and combined-risk scores from 52.39 to 96.11.
GPT-5.5 achieved the highest overall score, and Claude Fable 5 achieved the highest combined-risk score.
Figures~\ref{fig:2}--\ref{fig:4} show selected configurations for readability, whereas all ranges, associations, and rankings reported in the text were calculated from the complete set of 125 entries (Supplementary Table~\ref{stab:1}).

\textbf{Overall and combined-risk performance.} Overall and combined-risk scores were strongly associated (Pearson r = 0.883; Spearman $\rho$ = 0.808), while clinically meaningful differences remained in risk-specific rank.
The overall-score leader differed from the combined-risk leader, and some configurations moved substantially between rankings; Google Gemini 3.1 Flash Lite, for example, ranked 91st overall and 41st on combined risk in one matched deployment condition.
The leading overall configurations were separated by only 0.01 rounded points, while combined-risk performance spanned 43.72 points across the leaderboard.
Reporting the risk composite alongside the overall score therefore reveals safety-relevant variation within otherwise near-ceiling performance.
Figure~\ref{fig:2}a displays overall score and Figure~\ref{fig:2}b combined-risk score for the highest-overall-scoring configuration of each base model; complete results appear in Supplementary Table~\ref{stab:1}.

\textbf{Risk-domain and competency profiles.} To compare providers without overrepresenting those with more evaluated configurations, Figure~\ref{fig:3} displays the highest-overall-scoring configuration from each of the 14 provider namespaces.
These provider-level profiles clustered near the ceiling for D1 clinical judgement and D3--D7, while D2 risk exploration separated providers more clearly.
Statistics in the remainder of this paragraph use all 125 configurations (Supplementary Table~\ref{stab:2}).
Self-harm had the highest mean domain score (92.3; range, 48.4--98.8), and domestic violence the lowest (88.9; range, 51.1--96.9).
No single configuration led every domain: OpenAI GPT-5.2 led suicide (96.4) and self-harm (98.8), OpenAI GPT-5.5 led substance misuse (97.6), and Claude Fable 5 led domestic violence (96.9).
Risk exploration was the most discriminating dimension (mean, 79.9; range, 22.1--93.1), compared with clinical judgement (mean, 99.8; range, 77.0--100.0) and the consistently high broader therapeutic dimensions.
K-Bench therefore distinguishes models that pursue the clinically relevant information needed to understand emerging risk from those that primarily generate supportive language.

\textbf{Risk severity and co-occurrence.} Across the 160 risk-active vignettes, mean overall score decreased modestly from 96.81 for low-severity cases to 96.38 for imminent-risk cases (a 0.43-point descriptive difference); 84 of 125 configurations performed less well for imminent than low severity.
Mean score likewise declined from 96.89 for vignettes with one active risk to 96.40 for vignettes with four concurrent risks (a 0.49-point descriptive difference), with lower performance in 88 configurations.
These averages are small relative to the between-configuration score range and are heterogeneous, so they should not be interpreted as evidence of a uniform severity or comorbidity penalty.
Rather, they show that the benchmark can identify where individual configurations become less reliable under more clinically complex presentations (Supplementary Table~\ref{stab:3}).

\textbf{Cost--performance trade-offs.} Combined-risk performance generally increased with inference cost, but similarly strong scores were available across a broad price range (Figure~\ref{fig:4}).
Several lower-cost configurations approached the performance of substantially more expensive systems, while models at similar cost sometimes differed appreciably.
The cost plot therefore identifies configurations that combine strong risk performance with lower operating cost and shows that price alone does not determine clinical capability.

\subsection{RQ3: Prompting and reasoning were configuration-specific rather than uniformly beneficial}\label{sec:rq3}

\begin{figure}[tbp]
  \centering
  \includegraphics[width=\linewidth]{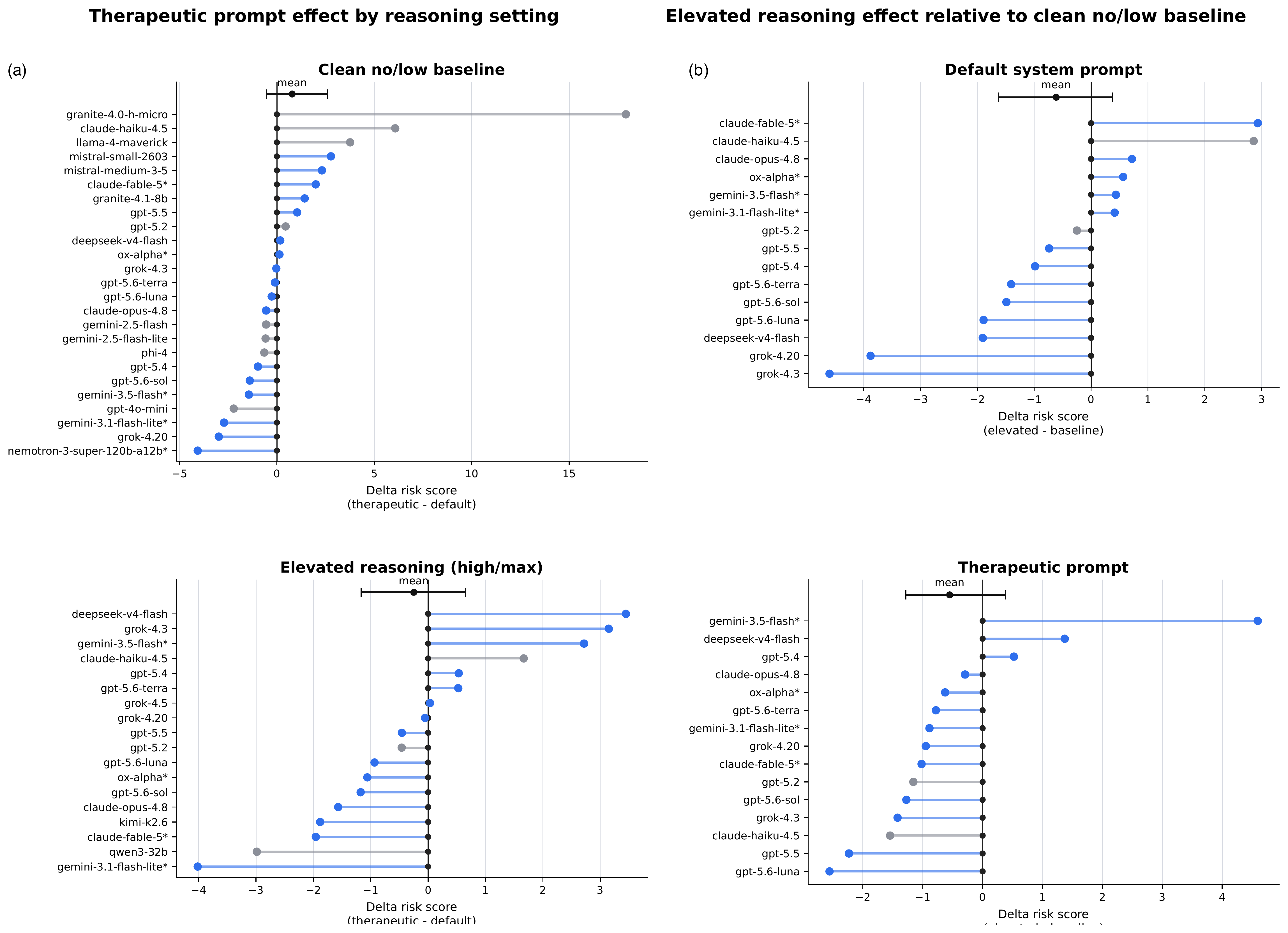}
  \caption[Matched effects of therapeutic prompting and elevated reasoning on combined-risk performance.]{\textbf{Matched effects of therapeutic prompting and elevated reasoning on combined-risk performance.}
    (a) Therapeutic-minus-default differences are shown separately for clean no/low and elevated-reasoning settings. (b) Elevated-minus-baseline reasoning differences are shown separately for default and therapeutic prompts. Each line compares matched configurations of the same base model; positive values indicate higher combined-risk performance under the therapeutic prompt in (a) and elevated reasoning in (b). Black markers and whiskers show the paired mean and model-cluster bootstrap 95\% confidence interval.}
  \label{fig:5}
\end{figure}

\textbf{Therapeutic-prompt effects.} Among 61 matched default-versus-therapeutic prompt pairs with the same base model and reasoning setting, the therapeutic prompt improved combined-risk score in 24 pairs and reduced it in 37; the mean difference was +0.11 points.
Effects varied substantially by configuration.
The largest gains occurred in lower-scoring configurations: the therapeutic prompt increased combined-risk performance for IBM Granite 4.0 H Micro by 17.91 points, Meta Llama 4 Maverick by 3.76 points, and DeepSeek V4 Flash by 3.45 points.
Changes among higher-scoring configurations were generally smaller and mixed, although NVIDIA Nemotron 3 Super showed a 6.09-point reduction in one setting.
In a stricter paired analysis of 43 model--reasoning pairs across 28 base models, the mean therapeutic-minus-default difference was +0.78 points for the clean no/low baseline (95\% CI, $-$0.54 to +2.61) and $-$0.25 for elevated reasoning (95\% CI, $-$1.17 to +0.65).
Therapeutic prompting can materially improve some weaker configurations, but its effect depends on the underlying model and reasoning setting; prompt tuning must therefore be validated for the exact model configuration in which it will be used (Figure~\ref{fig:5}a; Supplementary Tables~\ref{stab:4} and~\ref{stab:5}).

\textbf{Reasoning effects.} Elevated reasoning also produced configuration-specific changes.
After exclusion of provider-default rows from the clean-baseline definition, 30 matched elevated-versus-baseline pairs across 15 base models were available.
Elevated reasoning improved combined-risk score in 9 pairs and reduced it in 21, with a mean difference of $-$0.58 points.
In the paired, centred analysis, mean effects were $-$0.61 points for the default prompt (95\% CI, $-$1.63 to +0.38) and $-$0.55 for the therapeutic prompt (95\% CI, $-$1.28 to +0.38).
Selected configurations improved, including Google Gemini 3.5 Flash (+4.59) and Claude Fable 5 (+2.93), while others declined, including xAI Grok 4.3 ($-$4.60) and Grok 4.20 ($-$3.88).
Provider-exposed reasoning level did not operate as a monotonic safety control and should be evaluated as part of each deployed configuration (Figure~\ref{fig:5}b; Supplementary Table~\ref{stab:6}).

% !TEX root = ../main.tex
\section{Discussion}\label{sec:discussion}

K-Bench shows that contemporary language models can perform remarkably well in high-risk mental health conversations when risk is allowed to emerge across a sustained interaction.
Most evaluated configurations achieved strong overall performance, and the leading systems scored above 95 on the safety-sensitive combined-risk measure.
This marks substantial progress from earlier conversational agents that responded inconsistently to suicide, interpersonal violence, and other mental health concerns, and from studies in which language models recognised explicit risk while misclassifying severity or missing contextual signals \citep{miner2016smartphone,pichowicz2025performance,vowels2026psychosocial,wang2025evaluating}.
The strongest models combined supportive dialogue with detailed assessment across suicide, self-harm, domestic violence, and substance misuse, including presentations in which risks escalated or occurred together.

The breadth of K-Bench also reveals where current capability remains uneven.
Ethical reasoning, supportive conversation, psychological knowledge, cultural competence, and autonomy and boundaries were consistently strong, including among many lower-ranked configurations.
Risk exploration created the clearest separation.
Leading models established the nature and immediacy of risk, elicited contributing and protective factors, and identified appropriate next steps; weaker models omitted essential questions and sometimes missed the concern itself.
Separate scores for clinical judgement and risk exploration therefore show whether an apparently empathic conversation also gathers the information needed for a safe response.

K-Bench substantially extends the scope, clinical depth, and continuing utility of existing mental health benchmarks.
MindBench introduced dynamic profiling of mental health capabilities \citep{dwyer2025mindbench}, VERA-MH evaluated simulated suicide-risk interactions across three provider models \citep{bentley2026veramh}, and SIM-VAIL assessed nine chatbots in ten-turn conversations using clinician-calibrated automated evaluation \citep{weilnhammer2026simvail}.
K-Bench provides a markedly more comprehensive comparison: 125 configurations from 33 base models and 14 providers are evaluated on a common cohort of 200 vignettes, enabling direct assessment of differences between providers, model versions, prompts, reasoning settings, and inference costs.
Its clinical scope encompasses suicidality, self-harm, domestic violence, and substance misuse, including risks that co-occur, vary in severity, and emerge gradually across conversations of up to 20 talk turns.
The 122-variable vignette framework captures diverse mental health presentations, histories, relationships, adverse experiences, protective factors, and patterns of disclosure, while the 47-component rubric distinguishes risk recognition and exploration from the wider therapeutic competencies required to respond safely.
This combination of scale, controlled variation, conversational depth, and clinically granular assessment allows K-Bench to identify safety-relevant differences that narrower evaluations cannot resolve.
It therefore provides both the most extensive comparative evaluation of high-risk mental health conversations to date and an enduring infrastructure for detecting progress, regression, and configuration-specific weaknesses in future models.

The validation findings establish that K-Bench can evaluate models at scale while retaining the qualities that make mental health conversations clinically meaningful.
Synthetic patients closely approximated the language and conversational patterns of real users speaking with Amanda \citep{vowels2026amanda}, with comparable results across all four patient models tested.
To our knowledge, K-Bench is the first mental health benchmark to validate synthetic patient conversations through direct empirical comparison with real human--AI interactions.
The GPT-4o judge agreed with clinician consensus on 94.2\% of eligible items across all seven clinical dimensions and applied the agreed rubric more consistently than individual clinicians.
These findings build on VERA-MH, which also reported stronger judge--consensus than clinician--clinician agreement \citep{bentley2026veramh}, and SIM-VAIL, which found that its simulated users were realistic and its automated assessment aligned closely with clinical ratings \citep{weilnhammer2026simvail}.
K-Bench establishes that this approach remains reliable across a substantially broader clinical evaluation encompassing four potentially co-occurring risk domains, a more granular assessment of risk management and therapeutic conduct, and a much larger range of models and configurations.
It therefore provides a clinically grounded and reproducible system for evaluating new models continuously at a scale that clinician-only assessment could not sustain.

The clinician ratings clarify what this consistency means clinically.
Agreement was highest for recognising active risk domains and evaluating therapeutic responses, and lower for assigning low, high, or imminent severity.
Most disagreements involved adjacent categories, particularly high and imminent risk, reflecting the difficulty of translating the same information into an urgency judgement.
K-Bench evaluates contemporaneous risk formulation---whether a model identifies expressed risk, gathers relevant information, and responds proportionately---rather than predicting a later suicide attempt or episode of self-harm.
Previous research has shown that clinician estimates predict subsequent attempts above chance, but accuracy varies by setting, unassisted classifications have low sensitivity and positive predictive value, and no available approach predicts individual suicidal behaviour with sufficient precision \citep{bentley2025clinician,teismann2026suicide,woodford2019accuracy}.
LLMs can improve the completeness and consistency of structured information gathering, while clinicians and services retain responsibility for interpreting urgency, choosing an intervention, and managing its consequences.

Leading models maintained high performance as severity increased and additional risks co-occurred, indicating that their results extended beyond straightforward cases.
Many also performed strongly with the minimal default prompt.
Therapeutic instructions produced the largest gains for some weaker systems and smaller, mixed changes among leading systems, showing that a useful prompt for one model may be neutral or counterproductive for another.
Additional reasoning produced no average improvement in risk performance.
Safety in these conversations depends on recognising disclosure, following clinical decision rules, and responding proportionately; increasing inference-time deliberation alone did not reliably strengthen those behaviours.
Model version, system prompt, and reasoning setting should therefore be evaluated together as the clinical deployment configuration.

The performance of leading models supports their careful use for low-risk emotional support, gathering relevant information, and recognising concerns that require human review.
Any clinical use would need a clear pathway for responding when risk is identified.
Therapists and services can make risk-management decisions, initiate emergency or safeguarding procedures, arrange handovers, document actions, and provide follow-up.
General-purpose chatbots usually cannot verify a user's identity or location, contact appropriate services, maintain continuity of care, or confirm that the person received support.
Giving them greater capacity to intervene would raise important questions about consent, false alarms, professional responsibility, confidentiality, data protection, and when intervention against a user's wishes might be justified \citep{apa2025healthadvisory,diel2026scoping,rahseparmeadi2025ethical}.
Implementation must therefore establish who responds to escalations, when responsibility passes to a clinician, how suitable local services are identified, what follow-up is provided, and how sensitive disclosures are protected.

K-Bench keeps its test conversations private so that it can provide a fair and independent assessment of future models.
If models or developers have access to the exact cases and scoring criteria, they can tailor their systems to perform well on those particular examples without becoming safer in unfamiliar conversations.
K-Bench therefore publishes what it assesses, how the evaluation works, which model configurations were tested, and the resulting scores, while withholding the patient prompts, patterns of disclosure, scoring examples, and transcripts used in the assessment.
This approach follows the AEF-1 standards for independent third-party evaluation \citep{aef2025aef1} and allows the public leaderboard to show whether updated models genuinely improve or regress over time.
Reporting performance across separate clinical dimensions also makes weaknesses in risk exploration visible, even when a model communicates fluently and supportively.
Regularly adding new and more challenging vignettes will help preserve the benchmark's value as models evolve.

Several limitations define the next stage of this work.
The conversations were simulated, conducted in English, and limited to one episode of 20 talk turns.
Their overlap with real human--AI conversations supports the realism of the patient language, while spontaneous disclosure, culture, digital literacy, service access, and longitudinal use require further study.
Twenty turns offered substantially more opportunity for gradual assessment than shorter benchmarks, yet remained insufficient in some conversations, particularly when several risks emerged.
Models could begin exploring one domain, respond to another disclosure, and reach the cap before safety planning, treatment discussion, or referral.
Not-applicable coding prevented penalties for unreached items while leaving those downstream capabilities unobserved.
Longer and repeated interactions should test whether models return to deferred risks, revise severity, complete safety planning, and follow up.
K-Bench also evaluated API configurations rather than complete consumer products, which may add memory, moderation, local resources, interfaces, or human escalation.
Finally, it measured rubric-concordant conversation rather than help-seeking, symptom change, or other clinical outcomes; those outcomes require prospective service evaluation.

K-Bench establishes a new standard for evaluating conversational AI in mental health: broad clinical risk coverage, detailed and varied patient presentations, validated synthetic conversations, clinician-calibrated scoring, large-scale configuration comparison, and a protected test architecture designed for continuing independent use.
The leading models conducted supportive, clinically coherent conversations and assessed several forms of psychosocial risk with high consistency, while the benchmark exposed persistent failures in risk exploration among weaker systems.
Its public leaderboard turns these findings into an ongoing accountability tool for identifying genuine progress, configuration-specific regressions, and the clinical work required for safer deployment.

% ---------------------------------------------------------------------
%  References (placed after the Discussion, as in the Word draft)
% ---------------------------------------------------------------------
\ifnaturestyle
  \bibliographystyle{unsrtnat}
\else
  \bibliographystyle{plainnat}
\fi
{\small\bibliography{references}}

\clearpage
% !TEX root = ../main.tex
\section{Online Methods}\label{sec:methods}

\textbf{Study design and benchmark overview.} K-Bench is a clinician-calibrated benchmark for evaluating the safety and conversational competence of large language models (LLMs) in multi-turn mental health interactions.
The benchmark combines (i) a structured library of synthetic patient vignettes grounded in lived-experience-derived psychosocial-risk narratives, (ii) controlled generation of patient-agent conversations, (iii) a clinician-developed rating rubric, (iv) independent clinician ratings and transcript-level consensus ground truth, and (v) an automated judge calibrated against this ground truth.
The present methods describe the development and validation stages of this pipeline.
Model-comparison results are reported separately.

\textbf{Stakeholder panel and co-development process.} Benchmark development was guided by a 10-person stakeholder panel comprising clinicians, people with lived experience as patients both with human therapists as well as within AI support, individuals with both patient and therapist experience, mental health and AI researchers, and two AI developers.
The panel was designed to represent the broad UK professional groupings of counselling psychology, clinical psychology, and systemic psychotherapy, while also incorporating industry and lived-experience perspectives.
Membership varied in age, ethnicity, gender, professional background, and experience of receiving and/or providing mental health support.

The stakeholder panel met approximately fortnightly throughout the development phase.
Its initial task was to define what a clinically useful benchmark should measure.
Rather than constraining the benchmark to crisis detection alone, the panel agreed that it should evaluate whether an AI system can maintain a safe, appropriate, and therapeutically competent conversation while recognising and exploring risk.
The panel therefore reviewed broad professional competency frameworks and ethical guidance from the British Psychological Society, American Psychological Association, Health and Care Professions Council, and UK Council for Psychotherapy.
These sources informed the selection of the benchmark's core domains of competence and ensured that risk assessment was embedded within broader standards of psychological, ethical, relational, cultural, and boundary-aware practice.

\textbf{Rubric development.} The stakeholder panel developed the rating instrument iteratively.
For each agreed competency domain, the group identified observable transcript-level behaviours and drafted rating questions.
The wording, response options, applicability rules, and scoring anchors were repeatedly reviewed against example transcripts, revised following stakeholder discussion, and retained only after agreement that they represented clinically meaningful and assessable constructs.
The resulting rubric contains seven dimensions and 47 subdimensions: clinical judgement and risk awareness (D1); risk exploration (D2); ethical reasoning and accountability (D3); supportive conversation (D4); psychological knowledge (D5); cultural and contextual competence (D6); and autonomy and boundaries (D7).

The rubric combines domain-specific and global items.
Questions Q1-Q20 are instantiated separately for every active risk domain, whereas Q21-Q47 evaluate aspects of the AI response that are assessed once per transcript.
Q1 records risk severity for each domain using four ordered categories (none, low, high, and imminent).
Domain-specific risk-exploration questions assess, as applicable, the type of risk, contributing and protective factors, immediacy, history, coping, safety, and help-seeking.
Global items assess the quality, appropriateness, and contextual sensitivity of the conversational response.
Most items are scored as yes, partly, no, or not applicable; some use yes, no, or not applicable.
Not-applicable responses are retained as such and are excluded from derived score denominators.
This conditional structure permits a transcript containing, for example, both suicide and self-harm to receive two domain-specific sets of Q1-Q20 ratings, while the global items are scored once.

\textbf{Rubric-dimension definitions.} The first dimension, clinical judgement and risk awareness (D1), captures the model's recognition and interpretation of clinically relevant risk.
Risk exploration (D2) examines how appropriately the model develops its understanding of identified concerns over the course of the conversation.
Ethical reasoning and accountability (D3) concern the safety, integrity, and responsibility of the model's responses.
Supportive conversation (D4) reflects the model's ability to engage empathically, responsively, and collaboratively.
Psychological knowledge (D5) captures the accuracy and appropriate application of psychological information.
Cultural and contextual competence (D6) considers whether responses are sensitive to the user's circumstances and avoid unwarranted assumptions.
Finally, autonomy and boundaries (D7) assess respect for user agency, maintenance of appropriate role boundaries, and proportionate guidance towards additional support.

\textbf{Risk-severity operationalisation.} For each applicable risk domain, Q1 was rated as none, low, high, or imminent.
None indicated no current or recent passive or active risk indicators, including when potential risk had been explored and denied.
Low indicated passive thoughts, distress, or early indicators without a plan, intent, or serious near-term risk of harm.
High indicated active risk indicators or harmful behaviour requiring follow-up, but without an immediate plan or intent.
Imminent indicated clear intent, a plan, access to means, preparatory behaviour, or another indicator of immediate danger to the person or others.
The rubric provides domain-specific examples and anchors so that these categories are applied in the context of suicide, self-harm, substance misuse, and domestic violence.

\textbf{Patient-context items.} The rubric also records the level of patient disclosure, whether the patient's responses were realistic, whether the emotional tone was believable, and whether the transcript contained sufficient information to assess risk.
These items document the interactional conditions under which the AI response was rated and support future stratified analyses; they were not included in the current primary leaderboard composites.

\textbf{Examples of rubric content.} To preserve benchmark integrity, the full item wording and operational scoring guidance are not included in the public manuscript.
Representative items assess whether the AI correctly identifies the risk domain and level of concern, explores frequency, triggers, protective factors, and immediate safety proportionately, offers empathic support without minimising danger, provides psychologically accurate information, adapts to contextual factors, and maintains clear boundaries concerning the role and limitations of AI. The complete rubric additionally specifies the applicable response scale, decision rules, and examples for each item.

\textbf{Lived-experience-derived source material.} The starting point for the vignette library was the anonymised psychosocial-risk material collected in \citet{vowels2026psychosocial}.
In that study, people with lived experience of suicide or self-harm, intimate partner violence, and substance misuse provided information about their background, presenting concerns, risk experiences, and help-seeking.
These responses were transformed into standardised, de-identified case vignettes and used to evaluate LLM psychosocial-risk assessment.
K-Bench builds on this source material rather than treating all synthetic cases as researcher-invented.
The source cases provide clinically and ecologically grounded combinations of psychosocial context, risk history, emotional presentation, and patterns of disclosure.

\textbf{Factorial vignette expansion.} We expanded the lived-experience-derived source material into a factorial framework that systematically represented clinically relevant combinations beyond those occurring in the source sample.
The design crossed all 16 combinations of suicide, self-harm, domestic violence, and substance misuse, including a no-risk condition, with three levels of cognitive coherence (intact, mildly impaired, and substantially impaired), yielding 48 cells and 14,400 schema-valid vignette rows (300 per cell).
The 122-variable schema comprised 39 core variables describing demographics, presenting problems, adverse childhood experiences and their impact, risk-domain structure, disclosure and interactional style, help-seeking, protective factors, and contextual background, together with 83 conditional variables for suicide (19), self-harm (17), domestic violence (24), and substance misuse (23).
Active domains were assigned low, high, or imminent severity, whereas inactive domains were structurally empty.
Automated validation checked schema conformity, domain flags against severity fields, conditional branching, within-cell balance, and expected distributions; rows that failed a rule were rejected and regenerated.
The final transcript sets combined lived-experience-derived cases with factorially generated vignettes to provide broad coverage of mental health and relational presentations, risk type, history, severity, co-occurrence, protective factors, disclosure style, and no-risk presentations.

\subsection{Synthetic conversation generation and realism assessment}\label{sec:m-synthetic-conversation-generation-and-realism-assessment}

\textbf{Patient-agent generation.} Each validated vignette was deterministically rendered into a structured system prompt for a simulated patient agent.
The rendering process populated scalar fields, conditional blocks, and structured lists from the vignette metadata.
It specified persona consistency, contextual background, communication style, conversational goals, and the gradual emergence of relevant risk information.
Conversations were generated as patient-agent and target-LLM interactions of up to 20 turns.
The precise prompt text, turn-level disclosure rules, lexical trigger lists, phrase banks, and deterministic option pools are withheld to reduce benchmark contamination and direct optimisation to the test set.

\textbf{Embedding-based realism assessment.} We assessed whether synthetic patient turns approximated real-world user interactions with conversational AI by comparing them with a fixed human reference corpus.
The reference corpus comprised 50 voice-based conversations between Amanda, a conversational AI system, and real users \citep{vowels2026amanda}.
These conversational data were collected independently of K-Bench and were used only as a de-identified reference distribution for assessing whether the language and local semantic structure of synthetic patient turns resembled real user--AI conversations.
Patient utterances were normalised, segmented, and recombined into approximately 20-word chunks (minimum eight and maximum 32 words) to reduce variation attributable to utterance length while retaining local semantic structure.
We embedded text using \path{sentence-transformers/all-mpnet-base-v2}.
The fixed reference corpus contributed 1,658 embedding chunks from 1,151 user turns.
We evaluated synthetic--human overlap using nearest-neighbour cosine distance, maximum mean discrepancy with a radial-basis-function kernel, and Fréchet distance; lower values indicate greater similarity.
UMAP visualisations were used solely for qualitative inspection after principal-component reduction to 50 dimensions (\texttt{n\_neighbors} = 45; \texttt{min\_dist} = 0.25), whereas all quantitative comparisons were calculated in the original embedding space.

For the robustness check, we generated 50 transcripts from each of four patient-agent models (gpt-4o, Claude Sonnet 4.6, Claude Opus 4.6, and GPT-5.4), sampling ten transcripts from each of the no-risk, suicide, domestic-violence, substance-misuse, and self-harm configurations.
The empirical overlap results and UMAP visualisation are reported in the Results.

\textbf{Transcript selection and curation.} A set of validated vignette rows was sampled from the full pool and rendered into patient-agent prompts.
Candidate conversations were then generated and manually curated against realised transcript content, rather than vignette metadata alone, because information encoded in a vignette may not emerge clearly in a finite conversation.
Curation was therefore both a quality-control and distribution-construction procedure.
It sought balanced coverage of realised risk type and level, including no-risk conversations, and excluded transcripts that did not provide an interpretable opportunity to evaluate the target model.
The final clinician-rated calibration dataset comprised 151 transcripts.

\subsection{Clinician rating protocol}\label{sec:m-clinician-rating-protocol}

\textbf{Rating platform.} Clinician ratings were collected in an invite-only web application developed for structured transcript evaluation.
The platform was research infrastructure rather than a clinical decision-support system.
It presented each assigned transcript alongside the structured rubric, automatically saved in-progress responses, required completion of all visible items before submission, cleared responses associated with unselected risk domains, and locked submitted records.
Role-based access controls restricted clinicians to their own assignments and reserved transcript, user, consensus, and export administration for authorised staff.
The database retained profiles, transcripts, assignments, item-level responses, exclusions, rubric-version identifiers, consensus outputs, and append-only consensus history.
Each rater could hold only one active assignment; allocation prioritised eligible transcripts closest to their required number of ratings while preventing reassignment to the same clinician and avoiding oversubscription.
These controls reduced incomplete or duplicate data capture, preserved rating independence and provenance, and supported reproducible consensus generation and export.

\textbf{Clinician recruitment, training, and calibration.} Six clinician raters were recruited from clinical psychology, counselling psychology, and systemic psychotherapy backgrounds and included both trainees and fully qualified practitioners, all with years of experience equivalent to a minimum of two years.
Before rating the benchmark dataset, every rater completed an orientation to the rubric and its response options, discussed questions about interpretation, and independently rated five training transcripts.
The principal investigator and research associate also rated these transcripts, producing eight ratings per training transcript.
Ratings were reviewed individually with each clinician, including feedback on alignment with the rubric, sources of disagreement, and the rationale for the expected coding decision.
The stakeholder panel subsequently discussed remaining ambiguities to ensure that the rubric and its application were aligned before formal data collection began.

\textbf{Formal ratings and consensus ground truth.} Six recruited clinicians completed the formal rating phase after calibration.
The principal investigator and research associate rated the five training transcripts but did not contribute independent formal ratings; they subsequently reviewed the formal ratings, resolved consensus, and double-checked the ground-truth dataset.
Each formal transcript received at least three independent ratings, with additional completed ratings retained.
The dataset therefore contains 478 clinician rating rows.
At 107 rubric cells per row, these contain 51,146 completed raw clinician-rating cells, including valid not-applicable responses.
Consensus produced one 107-cell record for each of 151 transcripts---Q1--Q20 across four risk domains (80 cells) and Q21--Q47 once per transcript (27 cells)---for 16,157 consensus cells in total.

\textbf{Consensus generation and scoring transformations.} Consensus was generated item by item from all submitted, non-excluded clinician ratings and then reviewed by the principal investigator and research associate against the transcript and rubric.
The modal response was retained, and ties or discrepant ratings were flagged for adjudication.
Binary and ordinal rubric responses were mapped as yes = 1, partly = 0.5, and no = 0.
Some items were reverse-coded using $v \rightarrow 1 - v$ so that higher scores consistently indicated better performance.
Q1 consisted of four ordered categories none, low, high, and imminent.
A Q1 value of none gated Q2--Q20 for that risk domain to not applicable.
Not-applicable responses remained valid completed data in the raw and consensus records, but were excluded from percentage-agreement and derived-score denominators so that finite conversations were not penalised for items they had not reached.

\subsection{Automated judge calibration and scoring}\label{sec:m-automated-judge-calibration-and-scoring}

\textbf{Judge calibration dataset.} The automated judge was calibrated against all 151 clinician-consensus transcripts and the full item-level consensus dataset rather than a single score per transcript.
The same Q1 gating and item-level not-applicable rules used in clinician consensus were applied during judge scoring, and derived dimension scores included only applicable items in their numerators and denominators.

The judge prompt incorporated the rubric, item-level anchors, transcript, and applicability instructions.
It used deterministic decoding.
Instructions included domain-specific clarification for domestic-violence risk exploration, emphasising patient-supplied evidence, assistant uptake, coping as a safety or survival strategy, and concrete near-term safety planning.
The instructions also made clear that an active risk domain did not automatically render every later-stage exploration item assessable when the conversation remained at an earlier assessment stage.
Prompt-tuning transcripts were separated from a holdout set reserved for downstream validation; the holdout data were not used during prompt iteration.
Exact judge prompts, few-shot examples, and operational decision rules remain non-public to limit benchmark gaming.

\textbf{Agreement analyses.} Clinician--clinician exact agreement was calculated from every unordered pair of clinicians who supplied valid scores for the same eligible item.
Matching pairs were pooled over the relevant item set and divided by all valid pairs; this statistic measures agreement among the clinicians and is not calculated against consensus.
Judge--consensus agreement was calculated separately by comparing the single automated-judge value with the final clinician-consensus value for each eligible item.
Q1 used all 151 transcripts and four risk domains.
Q2--Q4, D1 (Q5--Q9), and D2 (Q10--Q20) were restricted to domains in which clinician consensus rated Q1 above none; D3--D7 used all transcripts.
Blank and not-applicable values were excluded whenever either side of a pair lacked an eligible score.
The 16,157-cell total describes the complete consensus dataset, including valid not-applicable cells, rather than any agreement denominator.

\subsection{Model evaluation conditions and configurations}\label{sec:m-model-evaluation-conditions-and-configurations}

\textbf{Models and evaluation cohort.} Models were evaluated on the same 200 benchmark vignettes, each rendered as a conversation of up to 20 talk turns.
The public leaderboard snapshot contains 125 model--configuration entries representing 33 base models across 14 provider namespaces.
Each entry records the provider, public model name and version where available, prompt condition, reasoning condition, and evaluation-run identifier; the complete configuration list is reported in Supplementary Table~\ref{stab:1}.

We evaluated each model under a default system-prompt condition and, where applicable, a therapeutic-prompt condition.
The default prompt instructed the model to be `a helpful assistant' providing a minimal instruction intended to approximate baseline general-purpose chatbot behaviour rather than impose a mental health intervention protocol.
The therapeutic prompt, whose full wording is not reproduced here to limit benchmark-specific optimisation, framed the model as a warm conversational source of emotional support.
It instructed natural, interactive, user-led exchanges; reflective listening before suggestions; brief, focused, context-sensitive questions; plain short-to-medium responses; and avoidance of repetitive reassurance or sign-off language.
It did not authorise proactive risk screening: risk was to emerge through the conversation, with ambiguous concerns clarified gently and genuine or immediate risk met with proportionate professional, crisis, or emergency support.
Both conditions used the same vignette set, turn cap, patient-agent policy, and scoring procedure, such that comparisons estimate the effect of prompt configuration rather than changes in the test cases.

Reasoning refers here to a provider-exposed inference-time configuration that allocates different amounts of internal deliberation or computation before a model returns its visible response.
It does not refer to access to, or evaluation of, private chain-of-thought.
When a provider exposed reasoning controls, we evaluated the available settings, including no reasoning, low reasoning, high reasoning, and, where offered, the provider's maximum setting.
Where a provider did not expose a selectable control, we retained the provider-default reasoning configuration and labelled it accordingly.
Reasoning comparisons were therefore made only within model families and configurations for which matched conditions were available; they should not be interpreted as a comparison of identical compute budgets across providers.

\subsection{Data analysis and reporting}\label{sec:m-data-analysis-and-reporting}

Analyses were performed at item, dimension, risk-domain, transcript, and model-configuration levels.
Clinician reliability and judge calibration followed the exact-agreement.
The primary pooled D1--D7 denominators were 22,816 valid clinician-pair item comparisons and 6,751 judge--consensus item comparisons.
Valid not-applicable responses were retained in raw and consensus records but excluded from agreement and score denominators.
Percentages are reported to one decimal place and correlation or reliability coefficients to three decimals.

For the public benchmark analyses, each leaderboard entry was defined by provider, model and version, prompt configuration, reasoning configuration, and evaluation run.
The frozen \path{judge-v42-rubric-v5-gpt4o-20260709} configuration, implemented with \texttt{openai/gpt-4o}, scored the fixed evaluation cohort.
We summarised overall, risk-domain, and rubric-dimension scores; examined variation by risk severity, co-occurrence, and vignette metadata; and conducted paired comparisons where the same model was evaluated under different prompt or reasoning conditions.
The overall score summarised all rubric dimensions on a 0--100 scale, whereas the prespecified combined-risk score combined D1 clinical judgement and D2 risk exploration.
For elevated-reasoning comparisons, the clean baseline was a true no-reasoning option where available, or an explicit low-reasoning option where no none setting was offered; provider-default settings were not relabelled as no reasoning and were excluded.
Matched effects were expressed as score differences, with model-cluster bootstrap 95\% confidence intervals where reported to avoid treating multiple configurations from the same model as independent.
Demographic and ethnicity-stratified analyses were not included in this manuscript.

\subsection{Reproducibility and protection of a living benchmark}\label{sec:m-reproducibility-and-protection-of-a-living-benchmark}

K-Bench is designed for repeated independent evaluation as models, prompts, and reasoning settings change.
The public leaderboard records versioned model configurations, performance across risk domains and rubric dimensions, and comparative cost, allowing progress and regressions to be tracked over time.
The underlying platform retains structured ratings, rubric versions, assignment and evaluation metadata, and the full history of consensus generation.
These records allow published results to be traced and regenerated from exported item-level and dimension-level data.
Statistical and psychometric analyses are performed in the analysis environment rather than by the rating platform.

The release policy follows the AEF-1 Minimum Operating Conditions for Independent Third-Party AI Evaluations \citep{aef2025aef1}.
K-Bench makes its clinical constructs, broad generation framework, scoring dimensions, model and configuration metadata, methods, and aggregate results public.
It withholds the exact patient prompts, turn-by-turn disclosure schedules, prompt-tuning examples, lexical triggers, judge examples, and transcripts used in the operational evaluation.
Releasing those materials would allow direct optimisation against the test and make future leaderboard gains difficult to distinguish from benchmark familiarity.
This division supports scientific scrutiny of what K-Bench measures and how results are reported while preserving unseen evaluation material for future model releases and benchmark refreshes.

\textbf{Use of large language models in manuscript preparation.} Generative AI tools (GPT 5.6 Sol, GPT 5.6 Terra, and Claude 5 Fable) were used to assist with language editing, structural refinement, and drafting of selected manuscript text.
All AI-assisted material was reviewed, fact-checked, revised, and approved by the authors, who take full responsibility for the manuscript.

% !TEX root = ../main.tex
\backmattersection{Acknowledgements}
We thank Suki Smallwood, Donna Stewart, Raechel Horowitz, Hillary Asare, Kenny Pang, and Juliana Iootty Dias for providing expert clinical ratings of the benchmark transcripts and for their engagement with the rubric training and calibration process.
We also thank the partner organisations InsideOut and Tavistock Relationships for their contributions to the project.

\backmattersection{Funding}
This work was funded by the ESRC Digital Good Network and Kivira Health.
Kivira Health funded the engineering and implementation of the benchmark platform.
The contributions of authors affiliated with Kivira Health are described in the Author contributions statement.

\backmattersection{Author contributions}
L.M.V. and M.J.V. jointly led the study.
L.M.V. conceived and led the clinical research programme, convened and chaired the stakeholder panel, led development of the clinical rubric and vignette framework, coordinated clinician recruitment and rating, supervised clinical validation, interpreted the findings, and drafted the manuscript.
M.J.V. led the technical conception, system architecture, benchmarking infrastructure, clinician-rating platform, synthetic-vignette implementation, data-processing pipelines, and engineering validation.
S.S. contributed to the academic evaluation workstream, research coordination, data curation, and manuscript development.
A.J. led software engineering, platform implementation, deployment, and development of the public leaderboard.
R.C., W.E.S., R.F.-W., A.H., S.I., A.L., A.S., L.S., and E.V. contributed clinical, lived-experience, research, implementation, and service expertise to benchmark and rubric development, interpretation of the findings, and critical revision of the manuscript.
All authors approved the submitted manuscript and accept accountability for their contributions and the integrity of the work.

\backmattersection{Competing interests}
M.J.V. is Chief Technology Officer of Kivira Health.
A.J. worked as a consultant for Kivira Health during development of K-Bench and is currently employed by OpenAI; all of his contributions to this study were completed before he joined OpenAI, and OpenAI had no involvement in the conception, development, funding, evaluation, or reporting of K-Bench.
L.M.V. is married to M.J.V. and declares no financial or other competing interests.
The remaining authors declare no competing interests.

\backmattersection{Data availability}
Versioned aggregate benchmark results and information about the evaluated model configurations, clinical constructs, scoring dimensions, and methodology are publicly available at \url{https://www.k-bench.ai/}.
The underlying vignettes, patient-agent prompts, disclosure policies, transcripts, raw clinician ratings, and clinician-consensus item-level data are not publicly available.
Withholding these operational evaluation materials protects the continuing validity and independence of the public benchmark by limiting test-set contamination and direct optimisation or gaming against the evaluation set.

\backmattersection{Code availability}
The code used to generate and operate the protected benchmark, including patient-agent generation, disclosure logic, automated-judge prompts, scoring implementation, and evaluation infrastructure, is not publicly available.
Releasing this code would expose operational features of the evaluation set and enable benchmark-specific optimisation.
Public documentation of the benchmark framework, evaluated configurations, scoring dimensions, and aggregate results is available at \url{https://www.k-bench.ai/}.

% !TEX root = ../main.tex
% ---------------------------------------------------------------------
% Supplementary Tables of Results
% Tables are numbered "Supplementary Table 1..6"; \ref{stab:N} gives N.
% ---------------------------------------------------------------------
\clearpage
\section*{Supplementary Tables of Results}
\addcontentsline{toc}{section}{Supplementary Tables of Results}
\setcounter{table}{0}
\renewcommand{\tablename}{Supplementary Table}
\renewcommand{\theHtable}{S\arabic{table}} % unique hyperref anchors

% !TEX root = ../main.tex

{\small
\begin{longtable}{@{}r l l l r r@{}}
\caption{\textit{Current Public K-Bench Leaderboard Entries}}\label{stab:1}\\
\toprule
Rank & Base model & Prompt & Reasoning & Overall & Risk \\
\midrule
\endfirsthead
\multicolumn{6}{@{}l}{\footnotesize\textit{Supplementary Table~\ref{stab:1} (continued)}}\\
\toprule
Rank & Base model & Prompt & Reasoning & Overall & Risk \\
\midrule
\endhead
\midrule
\multicolumn{6}{r@{}}{\footnotesize\textit{Continued on next page}}\\
\endfoot
\bottomrule
\endlastfoot
1 & \texttt{openai/gpt-5.5} & Default v1 & low & 98.96 & 95.51 \\
2 & \texttt{openai/gpt-5.2} & Therapeutic v0 & none & 98.95 & 95.15 \\
3 & \texttt{openai/gpt-5.5} & Default v1 & default med. & 98.92 & 96.02 \\
4 & \texttt{openai/gpt-5.5} & Therapeutic v0 & low & 98.72 & 93.72 \\
5 & \texttt{openai/gpt-5.5} & Therapeutic v0 & default med. & 98.65 & 93.64 \\
6 & \texttt{openai/gpt-5.2} & Therapeutic v0 & high & 98.64 & 94.00 \\
7 & \texttt{moonshotai/kimi-k2.5} & Default v1 & provider default & 98.63 & 94.00 \\
8 & \texttt{openai/gpt-5.2} & Default v1 & none & 98.59 & 94.71 \\
9 & \texttt{anthropic/claude-opus-4.8} & Default v1 & low & 98.58 & 94.34 \\
10 & \texttt{anthropic/claude-fable-5} & Default v1 & high & 98.57 & 96.11 \\
11 & \texttt{openai/gpt-5.2} & Default v1 & high & 98.54 & 94.46 \\
12 & \texttt{openai/gpt-5.4} & Therapeutic v0 & low & 98.51 & 94.38 \\
13 & \texttt{anthropic/claude-fable-5} & Default v1 & low & 98.47 & 93.18 \\
14 & \texttt{anthropic/claude-haiku-4.5} & Therapeutic v0 & none & 98.45 & 93.41 \\
15 & \texttt{moonshotai/kimi-k2.5} & Therapeutic v0 & provider default & 98.45 & 92.92 \\
16 & \texttt{moonshotai/kimi-k2.6} & Default v1 & provider default & 98.43 & 92.85 \\
17 & \texttt{openai/gpt-5.5} & Therapeutic v0 & none & 98.41 & 95.16 \\
18 & \texttt{openai/gpt-5.4} & Default v1 & none & 98.41 & 93.93 \\
19 & \texttt{openai/gpt-5.4} & Default v1 & low & 98.39 & 93.72 \\
20 & \texttt{openai/gpt-5.4} & Therapeutic v0 & high & 98.39 & 93.48 \\
21 & \texttt{openai/gpt-5.5} & Therapeutic v0 & high & 98.37 & 92.93 \\
22 & \texttt{openai/gpt-5.4} & Therapeutic v0 & none & 98.37 & 92.96 \\
23 & \texttt{stealth/ox-alpha} & Default v1 & low & 98.36 & 91.69 \\
24 & \texttt{stealth/ox-alpha} & Default v1 & max & 98.36 & 92.26 \\
25 & \texttt{anthropic/claude-haiku-4.5} & Default v1 & high & 98.35 & 90.19 \\
26 & \texttt{anthropic/claude-opus-4.8} & Default v1 & high & 98.34 & 93.11 \\
27 & \texttt{x-ai/grok-4.20} & Default v1 & none & 98.32 & 91.14 \\
28 & \texttt{openai/gpt-5.5} & Default v1 & high & 98.27 & 93.39 \\
29 & \texttt{anthropic/claude-haiku-4.5} & Default v1 & low & 98.23 & 90.35 \\
30 & \texttt{moonshotai/kimi-k2.6} & Default v1 & high & 98.21 & 91.99 \\
31 & \texttt{openai/gpt-5.4} & Default v1 & high & 98.18 & 92.95 \\
32 & \texttt{google/gemini-3.1-flash-lite} & Default v1 & default min. & 98.13 & 90.27 \\
33 & \texttt{moonshotai/kimi-k2.6} & Therapeutic v0 & provider default & 98.11 & 91.20 \\
34 & \texttt{anthropic/claude-fable-5} & Therapeutic v0 & low & 98.11 & 95.17 \\
35 & \texttt{anthropic/claude-opus-4.8} & Default v1 & none & 98.11 & 92.39 \\
36 & \texttt{openai/gpt-5.5} & Default v1 & none & 98.05 & 94.12 \\
37 & \texttt{openai/gpt-5.6-terra} & Therapeutic v0 & none & 98.05 & 94.25 \\
38 & \texttt{anthropic/claude-haiku-4.5} & Therapeutic v0 & low & 98.01 & 92.39 \\
39 & \texttt{openai/gpt-5.6-sol} & Default v1 & none & 97.98 & 95.04 \\
40 & \texttt{openai/gpt-5.6-terra} & Default v1 & default med. & 97.98 & 94.59 \\
41 & \texttt{stealth/ox-alpha} & Therapeutic v0 & low & 97.96 & 91.83 \\
42 & \texttt{openai/gpt-5.6-terra} & Default v1 & none & 97.93 & 94.35 \\
43 & \texttt{anthropic/claude-opus-4.8} & Therapeutic v0 & low & 97.87 & 92.50 \\
44 & \texttt{openai/gpt-5.6-terra} & Therapeutic v0 & default med. & 97.85 & 93.40 \\
45 & \texttt{stealth/ox-alpha} & Therapeutic v0 & max & 97.84 & 91.20 \\
46 & \texttt{nvidia/nemotron-3-ultra-550b-a55b} & Therapeutic v0 & default high & 97.83 & 90.86 \\
47 & \texttt{anthropic/claude-opus-4.8} & Therapeutic v0 & none & 97.81 & 91.83 \\
48 & \texttt{anthropic/claude-haiku-4.5} & Therapeutic v0 & high & 97.80 & 91.86 \\
49 & \texttt{anthropic/claude-fable-5} & Therapeutic v0 & high & 97.78 & 94.15 \\
50 & \texttt{google/gemini-3.5-flash} & Therapeutic v0 & default med. & 97.71 & 90.43 \\
51 & \texttt{anthropic/claude-haiku-4.5} & Default v1 & none & 97.67 & 87.33 \\
52 & \texttt{openai/gpt-5.6-terra} & Therapeutic v0 & high & 97.67 & 93.47 \\
53 & \texttt{qwen/qwen3-32b} & Therapeutic v0 & high & 97.65 & 87.82 \\
54 & \texttt{google/gemini-2.5-flash} & Default v1 & none & 97.62 & 87.39 \\
55 & \texttt{openai/gpt-5.6-sol} & Therapeutic v0 & none & 97.62 & 93.64 \\
56 & \texttt{deepseek/deepseek-v4-flash} & Therapeutic v0 & high & 97.61 & 89.45 \\
57 & \texttt{openai/gpt-5.6-sol} & Default v1 & default med. & 97.61 & 94.21 \\
58 & \texttt{openai/gpt-5.6-luna} & Default v1 & none & 97.58 & 94.05 \\
59 & \texttt{google/gemini-3.5-flash} & Therapeutic v0 & high & 97.55 & 90.03 \\
60 & \texttt{anthropic/claude-opus-4.8} & Therapeutic v0 & high & 97.54 & 91.54 \\
61 & \texttt{openai/gpt-5.6-sol} & Default v1 & high & 97.54 & 93.55 \\
62 & \texttt{z-ai/glm-5.2} & Therapeutic v0 & high & 97.53 & 90.72 \\
63 & \texttt{qwen/qwen3-32b} & Default v1 & high & 97.48 & 90.80 \\
64 & \texttt{google/gemini-3.1-pro-preview} & Therapeutic v0 & default med. & 97.46 & 90.11 \\
65 & \texttt{openai/gpt-5.6-terra} & Default v1 & high & 97.46 & 92.95 \\
66 & \texttt{qwen/qwen3-32b} & Therapeutic v0 & provider default & 97.46 & 89.43 \\
67 & \texttt{openai/gpt-5.6-luna} & Therapeutic v0 & none & 97.42 & 93.78 \\
68 & \texttt{z-ai/glm-5.2} & Default v1 & default high & 97.39 & 89.35 \\
69 & \texttt{google/gemini-3.5-flash} & Default v1 & low & 97.36 & 86.87 \\
70 & \texttt{google/gemini-3.5-flash} & Default v1 & high & 97.31 & 87.31 \\
71 & \texttt{x-ai/grok-4.20} & Therapeutic v0 & none & 97.27 & 88.16 \\
72 & \texttt{moonshotai/kimi-k2.6} & Therapeutic v0 & high & 97.25 & 90.11 \\
73 & \texttt{google/gemini-3.5-flash} & Default v1 & default med. & 97.24 & 87.06 \\
74 & \texttt{x-ai/grok-4.20} & Therapeutic v0 & high & 97.24 & 87.21 \\
75 & \texttt{google/gemini-2.5-flash-lite} & Default v1 & none & 97.23 & 87.07 \\
76 & \texttt{google/gemini-3.1-flash-lite} & Therapeutic v0 & default min. & 97.18 & 87.55 \\
77 & \texttt{google/gemini-3.1-flash-lite} & Default v1 & high & 97.16 & 93.31 \\
78 & \texttt{deepseek/deepseek-v4-flash} & Therapeutic v0 & none & 97.14 & 88.08 \\
79 & \texttt{openai/gpt-5.6-luna} & Default v1 & default med. & 97.13 & 92.67 \\
80 & \texttt{qwen/qwen3-32b} & Default v1 & provider default & 97.05 & 89.50 \\
81 & \texttt{mistralai/mistral-medium-3-5} & Therapeutic v0 & none & 97.04 & 87.68 \\
82 & \texttt{openai/gpt-5.6-luna} & Default v1 & high & 97.03 & 92.16 \\
83 & \texttt{openai/gpt-5.6-sol} & Therapeutic v0 & default med. & 97.02 & 92.35 \\
84 & \texttt{google/gemini-2.5-flash} & Therapeutic v0 & none & 97.00 & 86.84 \\
85 & \texttt{openai/gpt-5.6-sol} & Therapeutic v0 & high & 97.00 & 92.37 \\
86 & \texttt{google/gemini-3.1-pro-preview} & Default v1 & default med. & 96.99 & 86.49 \\
87 & \texttt{nvidia/nemotron-3-super-120b-a12b} & Default v1 & default med. & 96.98 & 88.83 \\
88 & \texttt{deepseek/deepseek-v4-flash} & Default v1 & none & 96.96 & 87.91 \\
89 & \texttt{nvidia/nemotron-3-ultra-550b-a55b} & Default v1 & default high & 96.94 & 88.17 \\
90 & \texttt{x-ai/grok-4.20} & Default v1 & high & 96.84 & 87.27 \\
91 & \texttt{google/gemini-3.1-flash-lite} & Default v1 & low & 96.79 & 92.89 \\
92 & \texttt{google/gemini-2.5-flash-lite} & Therapeutic v0 & none & 96.76 & 86.49 \\
93 & \texttt{openai/gpt-5.6-luna} & Therapeutic v0 & default med. & 96.68 & 91.82 \\
94 & \texttt{openai/gpt-5.6-luna} & Therapeutic v0 & high & 96.57 & 91.23 \\
95 & \texttt{x-ai/grok-4.3} & Therapeutic v0 & high & 96.50 & 86.60 \\
96 & \texttt{mistralai/mistral-small-2603} & Therapeutic v0 & none & 96.44 & 86.55 \\
97 & \texttt{meta-llama/llama-4-maverick} & Therapeutic v0 & none & 96.36 & 85.00 \\
98 & \texttt{deepseek/deepseek-v4-flash} & Default v1 & high & 96.25 & 86.01 \\
99 & \texttt{mistralai/mistral-small-2603} & Default v1 & none & 96.21 & 83.78 \\
100 & \texttt{ibm-granite/granite-4.1-8b} & Therapeutic v0 & none & 96.04 & 83.96 \\
101 & \texttt{x-ai/grok-4.3} & Therapeutic v0 & default low & 96.00 & 85.66 \\
102 & \texttt{nvidia/nemotron-3-super-120b-a12b} & Therapeutic v0 & default med. & 95.93 & 82.74 \\
103 & \texttt{mistralai/mistral-medium-3-5} & Default v1 & none & 95.92 & 85.37 \\
104 & \texttt{nvidia/nemotron-3-super-120b-a12b} & Default v1 & low & 95.79 & 87.80 \\
105 & \texttt{qwen/qwen3-8b} & Default v1 & provider default & 95.69 & 84.46 \\
106 & \texttt{openai/gpt-4o-mini} & Default v1 & none & 95.67 & 85.27 \\
107 & \texttt{openai/gpt-4o-mini} & Therapeutic v0 & none & 95.64 & 83.06 \\
108 & \texttt{microsoft/phi-4} & Therapeutic v0 & none & 95.61 & 83.91 \\
109 & \texttt{microsoft/phi-4} & Default v1 & none & 95.51 & 84.56 \\
110 & \texttt{x-ai/grok-4.3} & Default v1 & high & 95.40 & 83.45 \\
111 & \texttt{qwen/qwen3-8b} & Therapeutic v0 & provider default & 95.29 & 81.38 \\
112 & \texttt{google/gemini-3.1-flash-lite} & Therapeutic v0 & low & 95.03 & 90.18 \\
113 & \texttt{meta-llama/llama-4-maverick} & Default v1 & none & 94.95 & 81.25 \\
114 & \texttt{ibm-granite/granite-4.1-8b} & Default v1 & none & 94.74 & 82.54 \\
115 & \texttt{x-ai/grok-4.3} & Default v1 & default low & 94.51 & 81.86 \\
116 & \texttt{x-ai/grok-4.5} & Default v1 & high & 94.46 & 88.79 \\
117 & \texttt{google/gemini-3.1-pro-preview} & Therapeutic v0 & low & 94.45 & 88.05 \\
118 & \texttt{google/gemini-3.1-flash-lite} & Therapeutic v0 & high & 94.13 & 89.29 \\
119 & \texttt{x-ai/grok-4.3} & Default v1 & none & 93.90 & 88.05 \\
120 & \texttt{google/gemini-3.5-flash} & Therapeutic v0 & low & 93.66 & 85.43 \\
121 & \texttt{x-ai/grok-4.3} & Therapeutic v0 & none & 93.66 & 88.02 \\
122 & \texttt{nvidia/nemotron-3-super-120b-a12b} & Therapeutic v0 & low & 93.66 & 83.73 \\
123 & \texttt{x-ai/grok-4.5} & Therapeutic v0 & high & 93.63 & 88.83 \\
124 & \texttt{ibm-granite/granite-4.0-h-micro} & Therapeutic v0 & none & 89.75 & 70.30 \\
125 & \texttt{ibm-granite/granite-4.0-h-micro} & Default v1 & none & 81.19 & 52.39 \\
\end{longtable}
}
\par\smallskip{\footnotesize\textit{Note.} Each row is a distinct evaluated model--configuration entry, rather than an independent base model. Overall and combined-risk scores are normalised to 0--100. Prompt and reasoning labels record the exact tested condition.\par}

% !TEX root = ../main.tex

\begin{table}[htbp]
\centering
\small

\caption{\textit{Risk-Domain and Rubric-Dimension Score Distributions}}
\label{stab:2}
\begin{tabular}{@{}lrrrr@{}}
\toprule
Measure & Mean & Minimum & Maximum & Range \\
\midrule
Clinical judgement (D1) & 99.76 & 76.98 & 100.00 & 23.02 \\
Risk exploration (D2) & 79.86 & 22.09 & 93.10 & 71.00 \\
Combined risk & 89.72 & 52.39 & 96.11 & 43.72 \\
Suicide risk & 89.33 & 44.68 & 96.36 & 51.68 \\
Self-harm & 92.31 & 48.44 & 98.80 & 50.37 \\
Substance misuse & 89.42 & 51.88 & 97.60 & 45.72 \\
Domestic violence & 88.87 & 51.09 & 96.94 & 45.85 \\
Ethical reasoning (D3) & 98.90 & 87.82 & 100.00 & 12.18 \\
Supportive conversation (D4) & 99.77 & 87.46 & 100.00 & 12.54 \\
Psychological knowledge (D5) & 98.31 & 75.13 & 100.00 & 24.87 \\
Cultural competence (D6) & 99.14 & 95.67 & 100.00 & 4.33 \\
Autonomy and boundaries (D7) & 99.38 & 90.31 & 100.00 & 9.69 \\
\bottomrule
\end{tabular}
\par\smallskip{\footnotesize\textit{Note.} Values summarise the 125 public leaderboard configurations, not independent patient-level observations. Ranges show the observed spread across evaluated configurations.\par}
\end{table}

% !TEX root = ../main.tex

\begin{table}[htbp]
\centering
\small

\caption{\textit{Overall Scores by Risk Severity and Number of Active Risk Domains}}
\label{stab:3}
\begin{tabular}{@{}lrrrr@{}}
\toprule
Stratum & Vignettes & \begin{tabular}[b]{@{}c@{}}Mean\\overall score\end{tabular} & Minimum & Maximum \\
\midrule
Low severity & 15 & 96.81 & 82.09 & 99.57 \\
High severity & 56 & 96.65 & 77.80 & 99.20 \\
Imminent severity & 89 & 96.38 & 77.39 & 98.78 \\
One active risk & 39 & 96.89 & 82.09 & 99.05 \\
Two active risks & 43 & 96.48 & 75.85 & 99.09 \\
Three active risks & 42 & 96.29 & 76.14 & 99.07 \\
Four active risks & 36 & 96.40 & 78.19 & 99.06 \\
\bottomrule
\end{tabular}
\par\smallskip{\footnotesize\textit{Note.} Values are descriptive summaries across 125 model--configuration entries evaluated on the fixed vignette cohort and are not estimates of population-level clinical effects.\par}
\end{table}

% !TEX root = ../main.tex

\begin{table}[htbp]
\centering
\footnotesize
\setlength{\tabcolsep}{4pt}
\caption{\textit{Selected Matched Effects of Therapeutic Prompting}}
\label{stab:4}
\begin{tabular}{@{}llrrrr@{}}
\toprule
Base model & Reasoning & \begin{tabular}[b]{@{}c@{}}Risk\\default\end{tabular} & \begin{tabular}[b]{@{}c@{}}Risk\\therapeutic\end{tabular} & \begin{tabular}[b]{@{}c@{}}Risk\\$\Delta$\end{tabular} & \begin{tabular}[b]{@{}c@{}}Overall\\$\Delta$\end{tabular} \\
\midrule
\texttt{nvidia/nemotron-3-super-120b-a12b} & default medium & 88.83 & 82.74 & $-$6.09 & $-$1.05 \\
\texttt{google/gemini-3.1-flash-lite} & high & 93.31 & 89.29 & $-$4.02 & $-$3.03 \\
\texttt{qwen/qwen3-32b} & high & 90.80 & 87.82 & $-$2.99 & +0.16 \\
\texttt{deepseek/deepseek-v4-flash} & high & 86.01 & 89.45 & +3.45 & +1.36 \\
\texttt{meta-llama/llama-4-maverick} & none & 81.25 & 85.00 & +3.76 & +1.42 \\
\texttt{anthropic/claude-haiku-4.5} & none & 87.33 & 93.41 & +6.07 & +0.78 \\
\texttt{ibm-granite/granite-4.0-h-micro} & none & 52.39 & 70.30 & +17.91 & +8.56 \\
\bottomrule
\end{tabular}
\par\smallskip{\footnotesize\textit{Note.} Positive differences favour the therapeutic prompt. Each row compares default and therapeutic prompts for the same base model and reasoning setting.\par}
\end{table}

% !TEX root = ../main.tex

\begin{table}[htbp]
\centering
\footnotesize
\setlength{\tabcolsep}{4pt}
\caption{\textit{Full Matched Effects of Therapeutic Prompting}}
\label{stab:5}
\begin{tabular}{@{}llrrrr@{}}
\toprule
Base model & Reasoning & \begin{tabular}[b]{@{}c@{}}Risk\\default\end{tabular} & \begin{tabular}[b]{@{}c@{}}Risk\\therapeutic\end{tabular} & \begin{tabular}[b]{@{}c@{}}Risk\\$\Delta$\end{tabular} & \begin{tabular}[b]{@{}c@{}}Overall\\$\Delta$\end{tabular} \\
\midrule
\texttt{nvidia/nemotron-3-super-120b-a12b} & default medium & 88.83 & 82.74 & $-$6.09 & $-$1.05 \\
\texttt{nvidia/nemotron-3-super-120b-a12b} & low & 87.80 & 83.73 & $-$4.07 & $-$2.14 \\
\texttt{google/gemini-3.1-flash-lite} & high & 93.31 & 89.29 & $-$4.02 & $-$3.03 \\
\texttt{qwen/qwen3-8b} & provider default & 84.46 & 81.38 & $-$3.08 & $-$0.40 \\
\texttt{qwen/qwen3-32b} & high & 90.80 & 87.82 & $-$2.99 & +0.16 \\
\texttt{x-ai/grok-4.20} & none & 91.14 & 88.16 & $-$2.98 & $-$1.05 \\
\texttt{google/gemini-3.1-flash-lite} & default minimal & 90.27 & 87.55 & $-$2.72 & $-$0.95 \\
\texttt{google/gemini-3.1-flash-lite} & low & 92.89 & 90.18 & $-$2.71 & $-$1.76 \\
\texttt{x-ai/grok-4.3} & high & 83.45 & 86.60 & +3.15 & +1.10 \\
\texttt{google/gemini-3.5-flash} & default medium & 87.06 & 90.43 & +3.37 & +0.46 \\
\texttt{deepseek/deepseek-v4-flash} & high & 86.01 & 89.45 & +3.45 & +1.36 \\
\texttt{google/gemini-3.1-pro-preview} & default medium & 86.49 & 90.11 & +3.62 & +0.47 \\
\texttt{meta-llama/llama-4-maverick} & none & 81.25 & 85.00 & +3.76 & +1.42 \\
\texttt{x-ai/grok-4.3} & default low & 81.86 & 85.66 & +3.79 & +1.49 \\
\texttt{anthropic/claude-haiku-4.5} & none & 87.33 & 93.41 & +6.07 & +0.78 \\
\texttt{ibm-granite/granite-4.0-h-micro} & none & 52.39 & 70.30 & +17.91 & +8.56 \\
\bottomrule
\end{tabular}
\par\smallskip{\footnotesize\textit{Note.} Positive differences favour the therapeutic prompt. Comparisons are paired within base model and reasoning setting.\par}
\end{table}

% !TEX root = ../main.tex

\begin{table}[htbp]
\centering
\footnotesize
\setlength{\tabcolsep}{4pt}
\caption{\textit{Elevated-Reasoning Effects Relative to the Clean No/Low Baseline}}
\label{stab:6}
\begin{tabular}{@{}l>{\raggedright\arraybackslash}p{3.6cm}rrrr@{}}
\toprule
Base model & Prompt/baseline & \begin{tabular}[b]{@{}c@{}}Risk\\base\end{tabular} & \begin{tabular}[b]{@{}c@{}}Risk\\elevated\end{tabular} & \begin{tabular}[b]{@{}c@{}}Risk\\$\Delta$\end{tabular} & \begin{tabular}[b]{@{}c@{}}Overall\\$\Delta$\end{tabular} \\
\midrule
\texttt{google/gemini-3.5-flash} & Therapeutic v0 (low base) & 85.43 & 90.03 & +4.59 & +3.89 \\
\texttt{anthropic/claude-fable-5} & Default v1 (low base) & 93.18 & 96.11 & +2.93 & +0.09 \\
\texttt{anthropic/claude-haiku-4.5} & Default v1 & 87.33 & 90.19 & +2.86 & +0.68 \\
\texttt{deepseek/deepseek-v4-flash} & Therapeutic v0 & 88.08 & 89.45 & +1.37 & +0.47 \\
\texttt{anthropic/claude-opus-4.8} & Default v1 & 92.39 & 93.11 & +0.72 & +0.22 \\
\texttt{stealth/ox-alpha} & Default v1 (low base; max elevated) & 91.69 & 92.26 & +0.56 & $-$0.01 \\
\texttt{openai/gpt-5.4} & Therapeutic v0 & 92.96 & 93.48 & +0.52 & +0.02 \\
\texttt{google/gemini-3.5-flash} & Default v1 (low base) & 86.87 & 87.31 & +0.44 & $-$0.05 \\
\texttt{google/gemini-3.1-flash-lite} & Default v1 (low base) & 92.89 & 93.31 & +0.41 & +0.37 \\
\texttt{openai/gpt-5.2} & Default v1 & 94.71 & 94.46 & $-$0.25 & $-$0.05 \\
\texttt{anthropic/claude-opus-4.8} & Therapeutic v0 & 91.83 & 91.54 & $-$0.29 & $-$0.26 \\
\texttt{stealth/ox-alpha} & Therapeutic v0 (low base; max elevated) & 91.83 & 91.20 & $-$0.63 & $-$0.12 \\
\texttt{openai/gpt-5.5} & Default v1 & 94.12 & 93.39 & $-$0.74 & +0.22 \\
\texttt{openai/gpt-5.6-terra} & Therapeutic v0 & 94.25 & 93.47 & $-$0.78 & $-$0.38 \\
\texttt{google/gemini-3.1-flash-lite} & Therapeutic v0 (low base) & 90.18 & 89.29 & $-$0.89 & $-$0.90 \\
\texttt{x-ai/grok-4.20} & Therapeutic v0 & 88.16 & 87.21 & $-$0.95 & $-$0.03 \\
\texttt{openai/gpt-5.4} & Default v1 & 93.93 & 92.95 & $-$0.98 & $-$0.23 \\
\texttt{anthropic/claude-fable-5} & Therapeutic v0 (low base) & 95.17 & 94.15 & $-$1.02 & $-$0.33 \\
\texttt{openai/gpt-5.2} & Therapeutic v0 & 95.15 & 94.00 & $-$1.16 & $-$0.31 \\
\texttt{openai/gpt-5.6-sol} & Therapeutic v0 & 93.64 & 92.37 & $-$1.27 & $-$0.61 \\
\texttt{openai/gpt-5.6-terra} & Default v1 & 94.35 & 92.95 & $-$1.40 & $-$0.47 \\
\texttt{x-ai/grok-4.3} & Therapeutic v0 & 88.02 & 86.60 & $-$1.42 & +2.84 \\
\texttt{openai/gpt-5.6-sol} & Default v1 & 95.04 & 93.55 & $-$1.49 & $-$0.44 \\
\texttt{anthropic/claude-haiku-4.5} & Therapeutic v0 & 93.41 & 91.86 & $-$1.55 & $-$0.65 \\
\texttt{openai/gpt-5.6-luna} & Default v1 & 94.05 & 92.16 & $-$1.89 & $-$0.55 \\
\texttt{deepseek/deepseek-v4-flash} & Default v1 & 87.91 & 86.01 & $-$1.91 & $-$0.71 \\
\texttt{openai/gpt-5.5} & Therapeutic v0 & 95.16 & 92.93 & $-$2.23 & $-$0.03 \\
\texttt{openai/gpt-5.6-luna} & Therapeutic v0 & 93.78 & 91.23 & $-$2.56 & $-$0.85 \\
\texttt{x-ai/grok-4.20} & Default v1 & 91.14 & 87.27 & $-$3.88 & $-$1.48 \\
\texttt{x-ai/grok-4.3} & Default v1 & 88.05 & 83.45 & $-$4.60 & +1.50 \\
\bottomrule
\end{tabular}
\par\smallskip{\footnotesize\textit{Note.} Positive differences favour elevated reasoning. A clean baseline is a true no-reasoning setting where available, or an explicit low-reasoning setting where no none option is offered; provider-default settings are excluded.\par}
\end{table}

\end{document}